\documentclass[10pt,twocolumn,letterpaper]{article}

\usepackage{cvpr}              
\usepackage{algorithm}
\usepackage{algpseudocode}
\usepackage{multirow}
\usepackage{makecell}
\usepackage{xcolor}
\usepackage{booktabs}
\usepackage{colortbl}
\usepackage{graphicx}
\usepackage{array}
\usepackage{colortbl,xcolor}%
\usepackage{tcolorbox}
\tcbuselibrary{breakable,skins}

\definecolor{cvprblue}{rgb}{0.21,0.49,0.74}
\usepackage[pagebackref,breaklinks,colorlinks,allcolors=cvprblue]{hyperref}

\def\confName{CVPR}

\title{ReBind: Multi-Reference Video Editing via Structured Instructions with \\ Explicit Reference Relationships}

\author{
Xinyu Liu$^{1,2*}$ \
Shihao Li$^{3,2}$ \
Weihong Lin$^{2}$ \
Xinlong Chen$^{4,2}$ \
Yang Shi$^{5,2}$ \
Yujin Han$^{6}$ \
Yiyang Cai$^{1}$ \\
Yanghao Wang$^{1}$ \
Ruibin Yuan$^{1}$ \
Yuanxing Zhang$^{2\dagger}$ \
Pengfei Wan$^{2}$ \
Wenhan Luo$^{1\dagger}$ \
Yike Guo$^{1}$ \\
\\
$^{1}$HKUST \quad
$^{2}$Kling Team \quad
$^{3}$NJU \quad
$^{4}$UCAS \quad
$^{5}$PKU \quad
$^{6}$HKU
}
\begin{document}

\maketitle
\renewcommand{\thefootnote}{$\dagger$}
\footnotetext{Corresponding authors.}
\renewcommand{\thefootnote}{*}
\footnotetext{This work was conducted during the author's internship at Kling Team.}
\renewcommand{\thefootnote}{\arabic{footnote}}
\begin{abstract}
Recent diffusion-based video generation models have made significant progress in multi-reference image-conditioned video editing. 
However, existing methods still struggle to coordinate information from multiple visual sources accurately.
We identify a critical deficiency in existing approaches. Existing editing instructions lack explicit reference relationships, and most multimodal large language models (MLLMs) cannot generate them reliably.
To address this problem, we propose \textbf{ReBind}, a systematic framework that introduces semantic instructions with embedded reference tokens as the intermediate representation for multi-reference image-conditioned video editing.
Our key insight is embedding reference tokens at semantic positions to eliminate ambiguity and establish precise bindings between visual attributes and their sources.
We develop \textbf{ReBind-Instruct}, a specialized MLLM that learns to establish explicit bindings between visual attributes and their reference sources through a two-stage progressive scheme for precise reference relationships.
We further develop \textbf{ReBind-Edit}, which enables lightweight adaptation of text-to-video models to coordinate multiple references by binding visual attributes to their designated sources.
Extensive experiments demonstrate that ReBind substantially outperforms general-purpose MLLMs in instruction quality and achieves state-of-the-art performance among open-source methods on reference image conditioned video editing. Our project webpage:
https://rebind-mrv2v.github.io/.
\end{abstract}

\section{Introduction}

Recent advances in diffusion-based video generation~\cite{ho2022video,peebles2023scalable,wang2025wan,yang2025cogvideox,hacohen2601ltx,kong2024hunyuanvideo,wei2026dreamvideo} have made photorealistic synthesis increasingly accessible. As these models mature, multi-reference image-conditioned video editing (MRVE) emerges as a natural next step, enabling complex edits that coordinate information from multiple visual sources simultaneously. Consider transforming a street scene by applying cyberpunk lighting from one reference while replacing a character with a robot from another. While recent video editing methods have made progress in processing multiple reference images, existing methods struggle to accurately coordinate information from multiple visual sources, often failing to properly incorporate reference content or producing results inconsistent with editing instructions.

We trace this limitation to the lack of explicit reference relationships in existing instructions, which general-purpose vision-language models cannot precisely generate. First, existing training instructions, while containing reference anchors, rely on distant semantic dependencies (e.g., ``replace the character with the person \textit{from reference image 2}''), making it difficult for video editors to establish precise spatial binding between visual attributes and their sources. This leads to systematic failures that intensify with increasing reference count. Second, improving instruction quality requires strong \textit{comparative perception}: the ability to distinguish visual attributes across multiple references and establish correct source-feature associations. However, recent work has shown that existing multimodal large language models (MLLMs) struggle with comparative video understanding~\cite{wu2025vidic}, with this challenge becoming more severe in multi-reference scenarios. The lack of an appropriate \textbf{intermediate representation} that explicitly captures reference attribution is a key factor limiting MRVE performance.
\begin{figure*}[t]
\centering
\includegraphics[width=\textwidth]{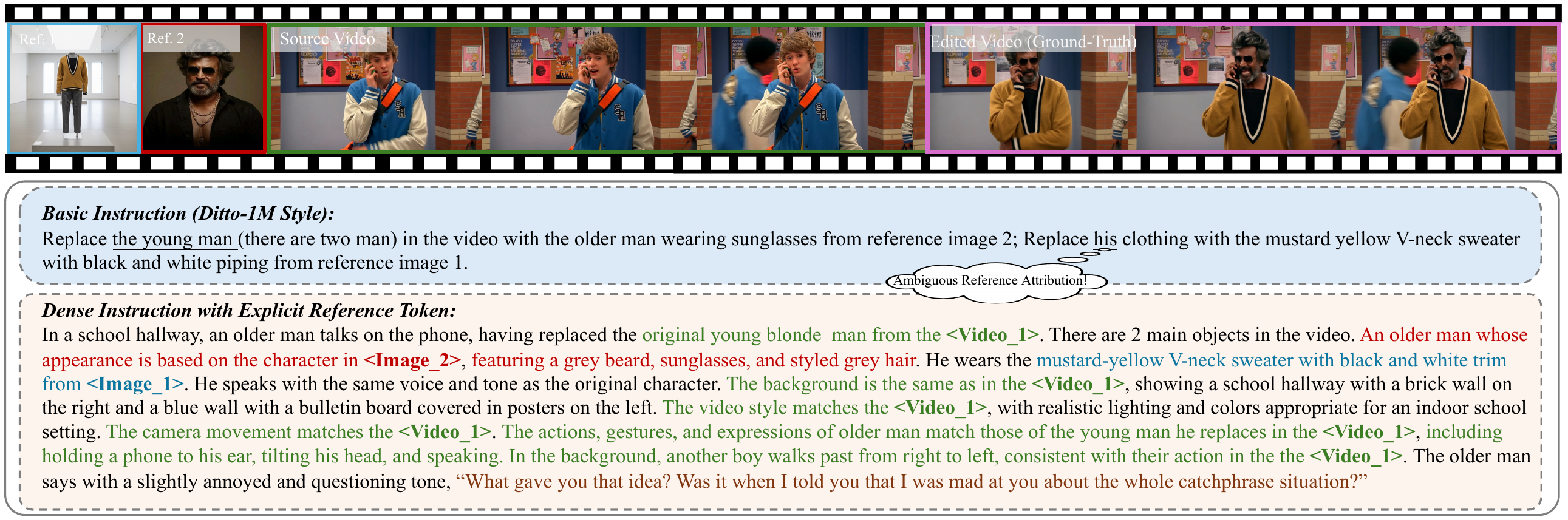}
\caption{Comparison between basic and dense instructions with explicit reference tokens, which are ground-truth training samples. In dense instructions, {\color{red!70!black}red}/{\color{cyan}blue} text indicates attributes from reference images 1/2, {\color{green!50!black}green} indicates preserved content, and {\color{brown}brown} describes audio content that provides semantic context to guide visual editing but remains unmodified.}
\label{fig:caption_case}
\vspace{-2mm}
\end{figure*}

To address this, we propose \textbf{ReBind}, a framework that introduces dense instructions with explicit reference relationships as the bridge between visual understanding and video generation. As illustrated in Figure~\ref{fig:caption_case}, the core design principle is embedding explicit visual reference tokens directly at their corresponding semantic positions in the instruction, eliminating ambiguity by clearly binding each visual attribute to its source. This design is inspired by recent advances in interleaved image-text modeling~\cite{zhang2026images}, which leverage the \textbf{contextual locality} of transformers for accurate multi-image understanding. However, generating such instructions with accurate reference attribution requires specialized capability that general-purpose MLLMs lack. We therefore develop \textbf{ReBind-Instruct}, a vision-language model trained through a two-stage pipeline: Structured Instruction Learning via supervised fine-tuning teaches the model to generate structured instructions with explicit visual reference tokens, while Reference Attribution Optimization employs reinforcement learning with tailored rewards to enhance reference attribution accuracy. Building on these dense instructions, we develop \textbf{ReBind-Edit} by extending the LTX-2.3 video generation model to support multi-reference conditioning via cross-attention mechanisms that bind visual features from reference images to their corresponding textual anchors, enabling precise multi-reference coordination.

To evaluate instruction generation quality, we augment the existing UniVBench~\cite{wei2026univbench} with human-annotated ground-truth dense instructions for both single- and multi-reference cases. On this testset, ReBind-Instruct substantially outperforms general-purpose MLLMs, demonstrating the effectiveness of our specialized training approach. For video editing quality, evaluation on IntelligentVBench~\cite{pan2026omniweaving} and UniVBench shows that ReBind-Edit achieves competitive performance with closed-source models and state-of-the-art results among open-source methods, with particularly strong performance on tasks requiring multi-reference coordination. Furthermore, our ablation studies confirm that the proposed dense instructions substantially outperform the short instructions widely adopted in prior work, and that higher-quality dense instructions consistently lead to better video editing performance.

Our contributions are summarized as follows:
\begin{itemize}[leftmargin=*, topsep=2pt, itemsep=1pt]
 \item We identify a critical deficiency in existing methods. Instead of dense captions, ReBind proposes a novel semantic instruction format that disambiguates reference relationships and enables models to better attend to pixel-level information.
    \item We develop ReBind-Instruct, a two-stage approach that addresses the fundamental weakness of existing MLLMs in discovering associations among multiple visual elements. Experiments demonstrate that ReBind-Instruct achieves near-parity with closed-source frontier MLLMs in reference attribution accuracy and intent recognition.
    \item Building upon ReBind-Instruct, ReBind-Edit enables lightweight adaptation of text-to-video models for multi-reference editing capabilities. ReBind-Edit substantially outperforms state-of-the-art open-source video editing models on UniVBench and IntelligentVBench benchmarks.
\end{itemize}

\section{Related Work}
\subsection{Video Captioning}
Current video captioning methods~\cite{zhang2023video,maaz2024video,shi2025mavors,sun2024video,chai2025auroracap} have made substantial progress in generating dense and accurate descriptions for individual videos. 
ShareGPT4Video~\cite{chen2024sharegpt4video} leverages GPT-4V to produce high-quality dense captions and improve video instruction data, while LLaVA-Video~\cite{zhang2024llava} constructs large-scale synthetic corpora to enhance video understanding. 
Recent studies further extend video captioning beyond visual-only descriptions: Qwen3-Omni~\cite{xu2025qwen3} supports joint audio-visual understanding with enhanced multimodal reasoning, and AVoCaDO~\cite{avocado} improves temporal alignment between audio and visual events. 
ViDiC~\cite{wu2025vidic} takes a step toward comparative video understanding by formulating video difference captioning over video pairs. 
However, existing methods do not address reference attribution in MRVE, where models must identify changes and ground each visual attribute to its corresponding reference source, requiring both comparative perception and explicit reference binding.

Reinforcement learning~\cite {rafailov2023direct,liu2026improving,shao2024deepseekmath} has proven effective for optimizing fine-grained video understanding beyond supervised fine-tuning. 
Video-SALMONN 2~\cite{tang2025video} employs direct preference optimization (DPO)~\cite{rafailov2023direct} to enhance audio-visual reasoning, while DiaDem~\cite{chen2026diadem} applies DPO for speaker attribution in dialogue captioning.
GRPO-based methods demonstrate effectiveness through carefully designed rewards. VideoCap-R1~\cite{meng2025videocap} optimizes action descriptions with structured reasoning rewards, and AVoCaDO~\cite{avocado} improves temporal alignment through checklist-based rewards.
Building on this paradigm, we apply GRPO with specialized rewards targeting reference attribution accuracy, a previously unaddressed objective in multi-reference editing.

\begin{figure*}[t]
\centering
\includegraphics[width=\textwidth]{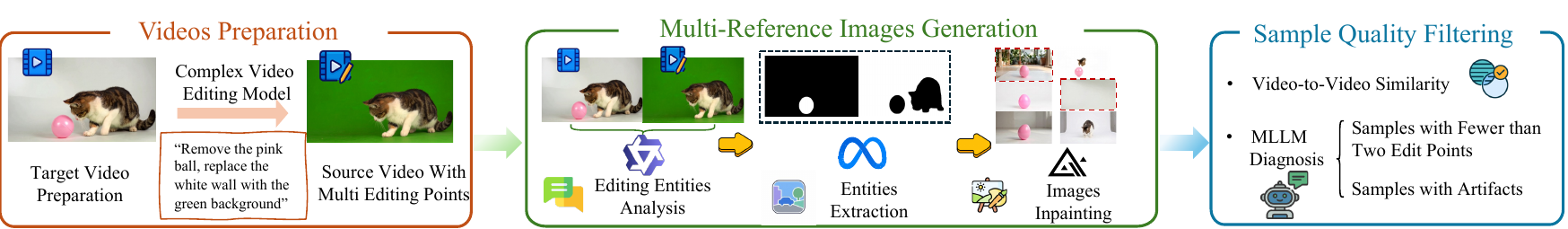}
\caption{Data construction pipeline for multi-reference video editing. We generate synthetic training triplets by applying inverse editing to target videos, followed by two-stage quality filtering and multi-reference image extraction.}
\label{fig:data_pipeline}
\end{figure*}
\subsection{Reference-Conditioned Video Editing}
Text-guided video editing~\cite{molad2023dreamix,zi2026senorita,he2025openve,wu2025insvie,qin2024instructvid2vid,chai2023stablevideo,yang2025videograin,wang2025videodirector,yu2025veggie,zhou2026omnishow} has enabled semantic manipulation of video content, but it remains limited in fine-grained control over appearance, identity, and localized visual attributes.
To improve controllability, reference-conditioned video editing methods~\cite{ye2025unic,liu2025stablev2v,jiang2025vace,wu2026loomvideo,liu2025revise,liu2026tide} introduce visual conditions to better preserve target appearance and editing fidelity.
More recently, unified video models~\cite{pan2025transfer,wei2025univideo,jin2025srum,chen2025blip3} integrate video understanding, generation, and editing within a single framework, further broadening the scope of reference-conditioned manipulation.
Notably, Bernini~\cite{liu2026bernini} unifies MLLMs and diffusion models through semantic planning in embedding space, achieving strong performance on video generation and editing with single-reference image-conditioned video editing (RVE) inputs.
However, multi-reference image-conditioned video editing (MRVE) remains an open challenge.
When generalized to MRVE tasks, they suffer from reference interference, ambiguous source-target correspondence, and inconsistent attribute extraction.

\section{Dataset Construction}
The multi-reference image-conditioned video editing (MRVE) task requires training samples consisting of multi-reference images, an input video, and a target video. We construct a synthetic dataset leveraging Bernini~\cite{liu2026bernini}, a text-guided complex video editing model. As illustrated in Figure~\ref{fig:data_pipeline}, our approach begins by collecting target videos from OpenVid-1M~\cite{nan2025openvid} and AVSET-10M~\cite{cheng2024avset}. We then apply the editing model to these target videos with inverse instructions, thereby generating the corresponding source videos through reverse editing operations. 

To ensure dataset quality, we employ a two-stage filtering process: (1) video-to-video cosine similarity to remove low-correspondence pairs, and (2) MLLM verification to eliminate incorrectly edited samples. For multi-reference image construction, we first utilize Qwen3-Omni-30B~\cite{bai2025qwen3} to identify and describe entities that undergo changes between source and target videos. We then apply SAM3~\cite{carion2025sam} to extract multi-frame object masks for each entity, followed by Flux2~\cite{labs2025flux} image inpainting to generate clean reference candidates. Finally, we select high-quality, front-facing images from each entity's candidate set as the final results. This pipeline yields approximately 100K video editing triplets with clean multi-reference images, consisting of approximately 50\% single-reference and 50\% multi-reference samples. These samples provide the visual data for our framework, with dense instruction annotations constructed in the corresponding training stages described below.
\section{Method}
Our framework consists of two key components: a dense instructor that generates structured instructions from video pairs and reference images, and a video editor that performs precise editing guided by these instructions. We train the dense instructor through a two-stage pipeline: (1) Structured Instruction Learning (Sec.~4.1) to learn the structured instruction format through supervised fine-tuning, and (2) Reference Attribution Optimization (Sec.~4.2) to optimize multiple reward objectives using reinforcement learning with Group Relative Policy Optimization (GRPO). The video editor (Sec.~4.3) is then trained on video editing triplets with dense instructions as text guidance.
\subsection{Stage 1: Structured Instruction Learning}
\label{sec:stage1}
We initialize our ReBind-Instruct from Qwen3-Omni-30B-A3~\cite{xu2025qwen3} and fine-tune it on a curated dataset of MRVE samples.

\noindent \textbf{Training Dataset.} To train the ReBind-Instruct, we select approximately 40K samples from the constructed triplets and annotate them with ground-truth dense instructions using Gemini 3.1 Pro. Each training instance consists of a source video $V_{src}$, an edited video $V_{edit}$, reference images $\{I_1, \ldots, I_K\}$, and a ground-truth dense instruction $S_{gt}$.
Crucially, these instructions contain explicit visual references using special tokens that establish precise bindings between textual descriptions and their corresponding visual sources.
 
To ensure annotation quality, we apply machine detection to filter erroneous data through automated quality checks: (1) reference token consistency verification to ensure all visual reference tokens are correctly formatted and match the provided images, and (2) rule-based filtering to detect length anomalies, repetitive content, and format errors. Detailed filtering prompts are provided in the Appendix.

\noindent \textbf{Training Objective.} We optimize the standard next-token prediction objective:
\begin{equation}
\label{eq:sft}
\mathcal{L}_{SFT} = -\sum_{t=1}^{T} \log P_\theta(s_t \mid s_{<t}, V_{src}, V_{edit}, I_1, \ldots, I_K)
\end{equation}
where $s_t$ denotes the $t$-th token in $S_{gt}$, $T$ is the sequence length, and $\theta$ represents the model parameters. This stage teaches the model to generate structured instructions with explicit visual references (e.g., \texttt{<Image\_1>}) at semantically appropriate positions by comparing the source and edited videos. After SFT, the model achieves strong capability in generating dense instructions for both single- and multi-reference image-conditioned video editing scenarios.

\subsection{Stage 2: Reference Attribution Optimization}
\label{sec:stage2}
\begin{figure*}[t]
\centering
\includegraphics[width=\textwidth]{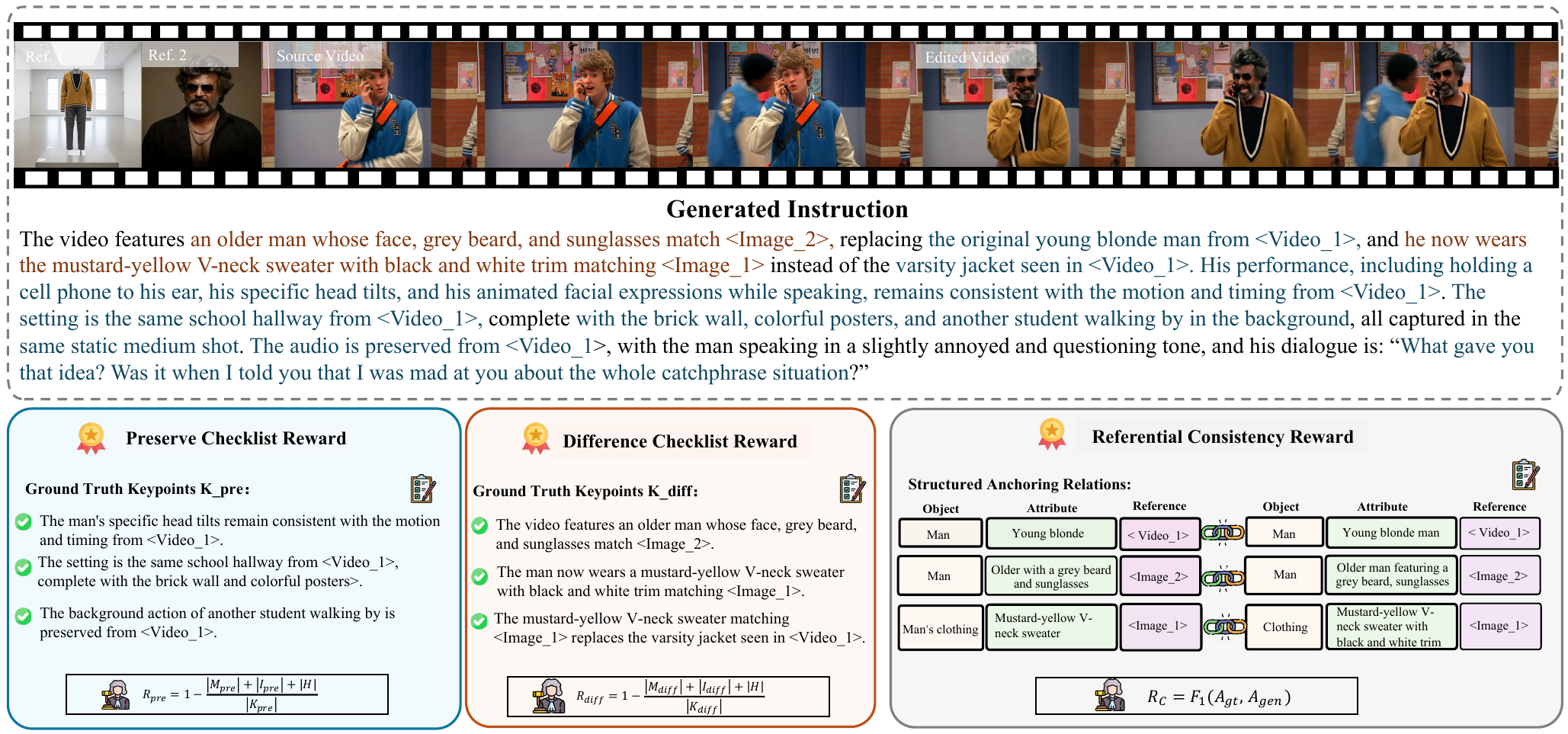}
\caption{Illustration of our dual checklist rewards for content completeness and referential consistency reward for accurate reference attribution. These two rewards, together with a length reward for mitigating repetition collapse, constitute our complete reward framework. }
\label{fig:reward_design}
\vspace{-3mm}
\end{figure*}
While SFT provides a strong initialization, it cannot directly optimize for task-specific objectives such as content completeness and reference grounding accuracy. To address this, we apply GRPO to further improve ReBind-Instruct by optimizing a composite reward function. We first describe the training data preparation, then introduce the GRPO algorithm and our reward design.

\noindent \textbf{Training Dataset.} For the Reference Attribution Optimization stage, we randomly sample approximately 2K video editing samples from the constructed dataset and manually annotate them with ground-truth dense instructions. To enable efficient reward evaluation during GRPO training, we perform offline extraction of reward components from human-annotated instructions using GPT-5.1, including dual checklist keypoints and relation triples.

\noindent \textbf{Group Relative Policy Optimization.} GRPO is a reinforcement learning algorithm that optimizes the policy by comparing outputs within groups. For each training instance $(V_{src}, V_{edit}, \{I_1, \ldots, I_K\}, S_{gt})$, we sample $G$ candidate instructions $\{S_1, \ldots, S_G\}$ from the current policy $\pi_\theta$. We then compute a reward $R(S_g)$ for each candidate and normalize rewards within the group:
\begin{equation}
\label{eq:grpo_advantage}
A_g = \frac{R(S_g) - \mu_R}{\sigma_R}
\end{equation}
where $\mu_R$ and $\sigma_R$ are the mean and standard deviation of rewards within the group. The policy is updated by maximizing:
\begin{equation}
\label{eq:grpo_objective}
\mathcal{L}_{GRPO} = \mathbb{E}_{S_g \sim \pi_\theta} \left[ A_g \cdot \log \pi_\theta(S_g) \right] - \beta \cdot D_{KL}(\pi_\theta \| \pi_{ref})
\end{equation}
where $\pi_{ref}$ is the reference policy (the SFT model), $D_{KL}$ is the KL divergence, and $\beta$ controls the KL penalty strength. This group-relative formulation provides more stable gradients and naturally handles reward scale variations.
\subsubsection{Reward Design}
To ensure both content completeness and reference grounding accuracy in MRVE instructions, we design a reward framework consisting of three core components: (1) dual checklist rewards $\mathcal{R}_{D}$ comprising $\mathcal{R}_{pre}$ and $\mathcal{R}_{diff}$ for content quality evaluation through missing and hallucination analysis, (2) referential consistency reward $\mathcal{R}_{C}$ for accurate reference attribution through structured triple matching, and (3) length reward $\mathcal{R}_{L}$ to mitigate repetition collapse and encourage appropriate instruction length.
\subsubsection{Dual Checklist Reward}
To enhance instruction completeness and accuracy, we propose dual checklist-based reward that evaluate content coverage through missing and hallucination analysis. The key motivation for dual rewards stems from the fundamental asymmetry in video editing captions: they must simultaneously describe \textit{what remains unchanged} and \textit{what gets modified}. These two aspects require distinct evaluation strategies due to their different semantic and structural requirements:

\textbf{Preserved content} refers to elements that maintain temporal continuity from the source video \texttt{<Video\_1>}. These descriptions focus on capturing the original dynamic or static features that persist through the editing process, such as ``the walking motion from \texttt{<Video\_1>}'' or ``the urban background preserved from the original scene''. The evaluation emphasizes whether the model correctly identifies and anchors unchanged content to the source video.

\textbf{Edited content} refers to elements that are modified by incorporating features from reference images \texttt{<Image\_X>}. These descriptions must explicitly attribute the modified features to their specific reference sources, such as ``the sunset sky from \texttt{<Image\_1>}''. The evaluation requires both identifying the modified content and ensuring correct reference attribution.

Each annotated ground-truth instruction $S_{gt}$ includes two distinct keypoint sets: preserved content keypoints $\mathcal{K}_{pre}$ and edited content keypoints $\mathcal{K}_{diff}$. We merge both sets into a unified keypoint set $\mathcal{K} = \mathcal{K}_{pre} \cup \mathcal{K}_{diff}$ for evaluation.

Given the unified keypoint set $\mathcal{K}$ and generated instruction $S_{gen}$, we employ GPT-5.1 to identify three types of events: (1) missing keypoints that exist in $\mathcal{K}$ but are absent in $S_{gen}$, (2) incorrectly described keypoints from $\mathcal{K}$ that are mentioned but inaccurately portrayed in $S_{gen}$, and (3) hallucinated content that appears in $S_{gen}$ but is absent from $\mathcal{K}$. For each identified missing or incorrect keypoint, we further categorize it as belonging to $\mathcal{K}_{pre}$ or $\mathcal{K}_{diff}$. Let $\mathcal{M}_{pre}$, $\mathcal{I}_{pre}$ denote missing and incorrect keypoints from preserved content, $\mathcal{M}_{diff}$, $\mathcal{I}_{diff}$ denote those from edited content, and $\mathcal{H}$ denote hallucinated keypoints. We compute separate rewards for each content type:
\begin{equation}
\mathcal{R}_{pre} = 1 - \frac{|\mathcal{M}_{pre}| + |\mathcal{I}_{pre}| + |\mathcal{H}|}{|\mathcal{K}_{pre}|}
\end{equation}
\begin{equation}
\mathcal{R}_{diff} = 1 - \frac{|\mathcal{M}_{diff}| + |\mathcal{I}_{diff}| + |\mathcal{H}|}{|\mathcal{K}_{diff}|}
\end{equation}
where $\mathcal{R}_{pre}$ evaluates preserved content description accuracy and $\mathcal{R}_{diff}$ evaluates edited content description with reference attribution. The final dual checklist reward is: $\mathcal{R}_{D} = \mathcal{R}_{pre} + \mathcal{R}_{diff}$.
\subsubsection{Referential Consistency Reward}
While checklist-based rewards evaluate content quality by verifying mentions of visual elements, they remain insufficient to assess whether these elements are correctly attributed to their source references. In reference-image coniditioned video editing scenarios, a particularly severe failure mode manifests: \textit{cross-reference confusion}, wherein the model accurately identifies visual content but assigns it to incorrect reference sources, thereby actively corrupting the editing process with plausible yet fundamentally incorrect directives.

To address this challenge, we introduce a referential consistency reward $\mathcal{R}_{cons}$ that prevents reference confusion at two levels through structured object-attribute-reference triples. Each anchoring relation is formulated as a triple $a_i = \langle o_i, d_i, r_i \rangle$ comprising an object identifier, visual attribute, and reference anchor. This design enforces correct attribution both within individual triples and across multiple triples to prevent cross-reference confusion. Ground-truth relations constitute the set $\mathcal{A}_{gt} = \{a_1^{gt}, \ldots, a_M^{gt}\}$ whilst generated relations form $\mathcal{A}_{gen} = \{a_1^{gen}, \ldots, a_N^{gen}\}$.

We employ GPT-5.1 to extract triples from both ground-truth and generated captions, then construct a binary matching matrix $W \in \{0,1\}^{M \times N}$ where each element $w_{i,j}$ indicates whether triple $a_i^{gt}$ matches triple $a_j^{gen}$:
\begin{equation}
\label{eq:triple_score}
w_{i,j} = \begin{cases}
1, & \text{if } o_i = o_j \text{ and } r_i = r_j \text{ and } d_i = d_j \\
0, & \text{otherwise}
\end{cases}
\end{equation}
A match requires exact agreement on all three components: object identity, reference anchor, and visual descriptor. This strict matching criterion eliminates cross-reference confusion by assigning zero score to cases where the object and descriptor are correct but anchored to the wrong reference.
We decompose the evaluation into recall and precision. Recall measures how well the generated instruction covers ground-truth content by finding, for each GT triple, whether any generated triple matches it:
\begin{equation}
\label{eq:ref_recall}
\text{Recall} = \frac{1}{M} \sum_{i=1}^{M} \max_{j=1}^{N} w_{i,j}
\end{equation}
Precision measures how accurate the generated content is by finding, for each generated triple, whether any GT triple matches it:
\begin{equation}
\label{eq:ref_precision}
\text{Precision} = \frac{1}{N} \sum_{j=1}^{N} \max_{i=1}^{M} w_{i,j}
\end{equation}
The referential consistency reward combines these components via F1-score:
\begin{equation}
\label{eq:ref_consistency_f1}
\mathcal{R}_{C} = \frac{2 \cdot \text{Precision} \cdot \text{Recall}}{\text{Precision} + \text{Recall}}
\end{equation}
\subsubsection{Length Reward}
To mitigate output repetition collapse and enhance inference efficiency, we design a length-regularized reward that encourages complete captions while penalizing excessive length. Following~\cite{avocado}, we adopt a unidirectional penalty structure. The thresholds $\tau_1 = 512$ and $\tau_2 = 1024$ are set based on empirical analysis, where $\tau_1$ is approximately the longest instruction in the ground-truth distribution and $\tau_2$ is approximately twice the maximum ground-truth length:
\begin{equation}
\label{eq:length}
\mathcal{R}_{L} = \begin{cases}
1.0, & \text{if } L < \tau_1 \\
1 - \frac{L - \tau_1}{\tau_2 - \tau_1}, & \text{if } \tau_1 \leq L < \tau_2 \\
0.0, & \text{otherwise}
\end{cases}
\end{equation}
where $L$ denotes the token count of the generated caption.

\subsection{Reference-Conditioned Video Editor}
\label{sec:editor}
After training ReBind-Instruct, we develop ReBind-Edit that performs precise video editing guided by dense instructions with explicit reference tokens. We build our editor upon LTX-2.3~\cite{hacohen2026ltx}, a representative text-to-video generation model.

\noindent \textbf{Architecture.} We build upon LTX-2.3~\cite{hacohen2026ltx}, a text-to-video diffusion transformer originally designed for text-guided generation. To enable multi-reference image-conditioned video editing, we extend the model to accept a concatenated latent sequence: $z = [z_{ref_1}, \ldots, z_{ref_K}, z_{src}, z_{edit}]$, where reference images and source video serve as clean conditioning, while the edited video undergoes the diffusion process~\cite{team2025kling}. The source video $V_{src}$ includes both visual and audio modalities, with audio providing semantic context to guide visual editing while remaining unmodified in the output. The model incorporates cross-attention layers that bind visual features from reference images to their corresponding textual anchors (\texttt{<Image\_X>}, \texttt{<Video\_X>}) in the dense instruction, enabling precise extraction of attributes from designated sources as directed by the instruction. This design allows the model to handle variable numbers of references within a unified architecture.
\begin{figure*}[t]
\centering
\includegraphics[width=\textwidth]{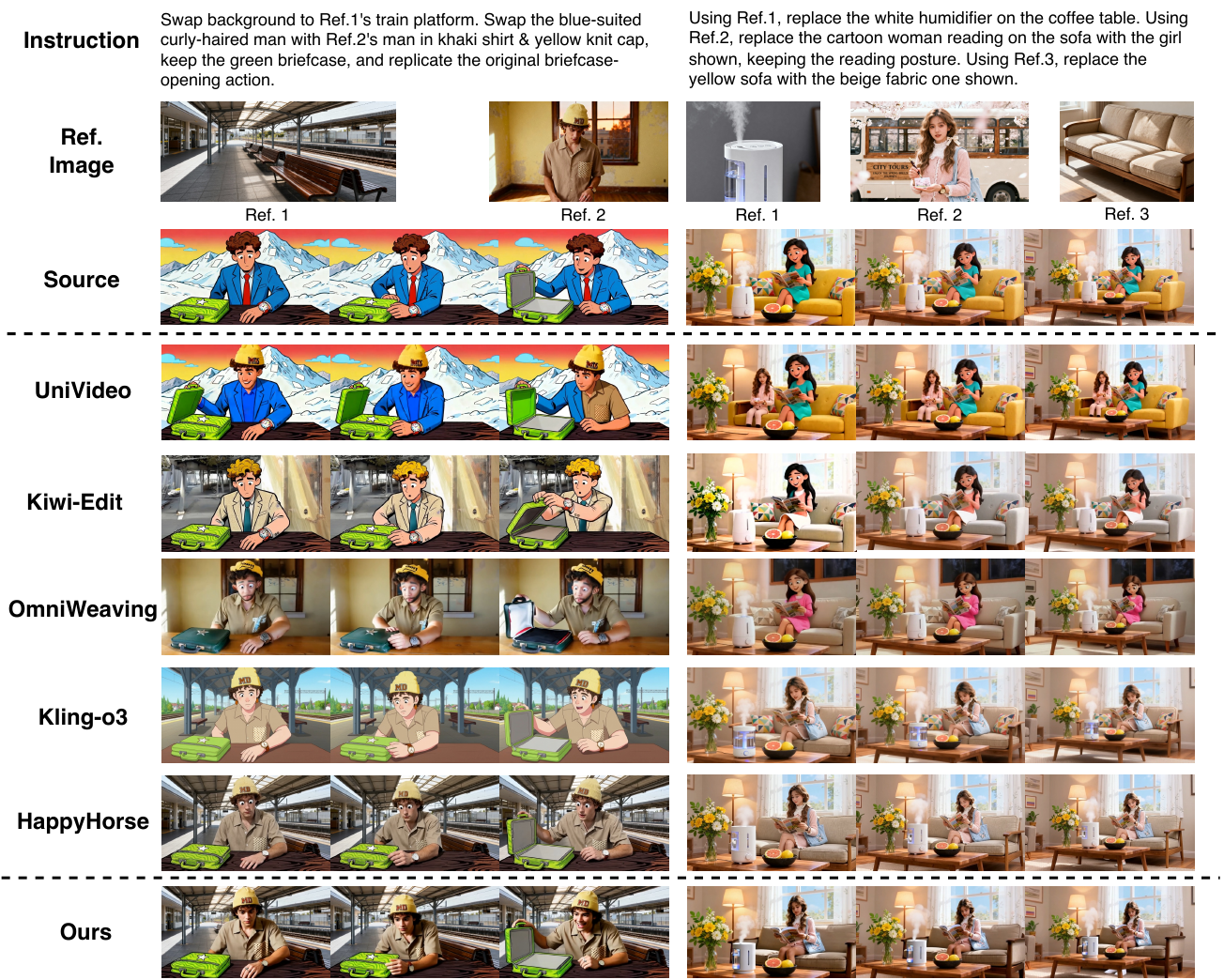}
\caption{Qualitative comparison results of multi-reference images conditioned video editing on UniVBench.}
\label{fig:qualitative_case}
\vspace{-2mm}
\end{figure*}

\noindent \textbf{Training.} We train the reference-image conditioned model on video editing samples $(V_{src}, V_{edit}, \{I_1, \ldots, I_K\}, S_{gt})$ where $V_{src}$ is the source video, $V_{edit}$ is the edited video, $\{I_1, \ldots, I_K\}$ are reference images, and $S_{gt}$ is the dense instruction generated by our trained dense captioner. We encode them into latents and construct the concatenated sequence $z = [z_{ref_1}, \ldots, z_{ref_K}, z_{src}, z_{edit}]$. During training, we apply noise only to the edited video portion $z_{edit}$, obtaining $z_{edit,t} = \sqrt{\bar{\alpha}_t} z_{edit} + \sqrt{1-\bar{\alpha}_t} \epsilon$ where $\epsilon \sim \mathcal{N}(0, I)$. The training objective is:
\begin{equation}
\label{eq:editor_unified}
\mathcal{L} = \mathbb{E}_{z,t,\epsilon} \left[ \| \epsilon - \epsilon_\theta([z_{ref_1}, \ldots, z_{ref_K}, z_{src}, z_{edit,t}], t, S_{gt}) \|^2 \right]
\end{equation}
where $\epsilon_\theta$ is the denoising network that processes the full concatenated sequence with cross-attention to the dense instruction $S_{gt}$.
\section{Experiments}
We conduct extensive experiments to validate the effectiveness of our approach across three key dimensions. First, we evaluate whether our specialized two-stage training produces higher-quality reference-grounded instructions compared to general-purpose vision-language models. Second, we investigate whether dense instructions with explicit reference attribution actually improve video editing quality over conventional short instructions through controlled ablation studies. Third, we benchmark our unified video editor against state-of-the-art methods on both single- and multi-reference editing tasks.
\begin{table}[t]
\centering
\caption{Comparison on the single-reference image-conditioned video editing (RVE) task evaluated on IntelligentVBench.}
\label{tab:editing_results}
\resizebox{\columnwidth}{!}{
\renewcommand{\arraystretch}{1.25}
\begin{tabular}{l|ccccc}
\toprule
\multirow{2}{*}{\textbf{Method}} & \multicolumn{5}{c}{\textbf{Single Reference}$\uparrow$} \\
\cmidrule(lr){2-6}
 & \textbf{IF} & \textbf{CP} & \textbf{VQ} & \textbf{Min} & \textbf{Avg} \\
\midrule
\multicolumn{6}{l}{\textit{Closed-Source Models}} \\
\textcolor{gray}{Kling-o3 (2026-03)~\cite{klingo3}} & \textcolor{gray}{4.60} & \textcolor{gray}{4.13} & \textcolor{gray}{3.91} & \textcolor{gray}{4.21} & \textcolor{gray}{3.60} \\
\textcolor{gray}{HappyHorse 1.0~\cite{happyhorse}} & \textcolor{gray}{2.93} & \textcolor{gray}{4.41} & \textcolor{gray}{3.89} & \textcolor{gray}{3.74} & \textcolor{gray}{2.70} \\
\midrule
\multicolumn{6}{l}{\textit{Open-Source Models}} \\
VACE-Wan2.1~\cite{jiang2025vace} & 1.46 & 1.42 & 1.71 & 1.27&1.53 \\
VACE-LTX~\cite{jiang2025vace} & 1.43 & 1.36 & 1.25 & 1.20  & 1.35 \\
VINO~\cite{chen2026vino} & 2.86 & 2.90 & 2.52 & 2.11 & 2.76\\
UniVideo(query)~\cite{wei2025univideo} & 3.22 & 3.91 & 3.26 & 2.66 & 3.46 \\
UniVideo(hidden)~\cite{wei2025univideo} & 3.13 & 4.01 & 2.93 & 2.56 & 3.36\\
OmniWeaving~\cite{pan2026omniweaving} & \underline{4.00} & \underline{4.04} & \underline{3.65} & \underline{3.31}& \underline{3.89}  \\
\midrule
\textbf{ReBind-Edit (Ours)} & \textbf{4.59} & \textbf{4.41} & \textbf{4.16} & \textbf{3.94} & \textbf{4.38} \\
\bottomrule
\end{tabular}
}
\end{table}
\begin{table*}[t]
\centering
\caption{Quantitative comparison on single-reference image-conditioned video editing (RVE) and multi-reference image-conditioned video editing (MRVE) evaluated on UniVBench. Scores range from 0-10.
\textbf{Bold}: best, \underline{underline}: second best.}
\label{tab:multref_comparison}
\resizebox{\textwidth}{!}{
\renewcommand{\arraystretch}{1.2}
\begin{tabular}{l|cccccccc|c|cccccccc|c}
\toprule
\multirow{2}{*}{\textbf{Method}} & \multicolumn{9}{c|}{\textbf{Single Reference}$\uparrow$} & \multicolumn{9}{c}{\textbf{Multi Reference}$\uparrow$} \\
\cmidrule(lr){2-10} \cmidrule(lr){11-19}
 & \textbf{Sub.} & \textbf{Back.} & \textbf{Action} & \textbf{Cam.} & \textbf{Color} & \textbf{Light.} & \textbf{Style} & \textbf{Rel.Pos.} & \textbf{Overall} & \textbf{Sub.} & \textbf{Back.} & \textbf{Action} & \textbf{Cam.} & \textbf{Color} & \textbf{Light.} & \textbf{Style} & \textbf{Rel.Pos.} & \textbf{Overall} \\
\midrule
\multicolumn{19}{l}{\textit{Closed-Source Models}} \\
\textcolor{gray}{Kling-o3 (2026-03)~\cite{klingo3}} & \textcolor{gray}{6.42} & \textcolor{gray}{7.94} & \textcolor{gray}{7.51} & \textcolor{gray}{8.61} & \textcolor{gray}{7.36} & \textcolor{gray}{7.85} & \textcolor{gray}{6.65} & \textcolor{gray}{7.65} & \textcolor{gray}{7.50} & \textcolor{gray}{5.99} & \textcolor{gray}{6.45} & \textcolor{gray}{6.84} & \textcolor{gray}{8.01} & \textcolor{gray}{6.96} & \textcolor{gray}{7.37} & \textcolor{gray}{5.85} & \textcolor{gray}{7.07} & \textcolor{gray}{6.82} \\
\midrule
\multicolumn{19}{l}{\textit{Open-Source Models}} \\
UniVideo(query)~\cite{wei2025univideo}  & 3.08 & 3.51 & 5.97 & 8.03 & 5.04 & 5.93 & 4.02 & 6.92 & 5.31 & \underline{4.61} & 4.75 & 5.90 & \textbf{7.80} & 5.73 & 6.22 & 4.83 & 6.39 & 5.78 \\
    UniVideo(hidden)~\cite{wei2025univideo}  & \underline{5.15} & \underline{6.82} & 6.02 & \underline{8.03} & \underline{6.36} & \underline{7.12} & \underline{5.58} & 6.76 & \underline{6.48} & 4.57 & \underline{4.84} & 5.67 & 7.64 & \underline{5.89} & \underline{6.44} & \underline{4.96} & 6.49 & \underline{5.81} \\
Kiwi-Edit~\cite{lin2026kiwi} & 3.14 & 3.38 & \underline{6.23} & 7.96 & 4.80 & 5.86 & 4.08 & \underline{7.07} & 5.32 & 4.18 & 4.67 & \underline{6.05} & \underline{7.79} & 5.63 & 6.25 & 4.81 & \underline{6.65} & 5.75 \\
OmniWeaving~\cite{pan2026omniweaving} & 3.08 & 2.95 & 6.04 & 7.67 & 4.22 & 5.46 & 3.68 & 6.82 & 4.99 & 4.28 & 4.00 & 5.65 & 7.34 & 5.32 & 5.76 & 4.29 & 6.02 & 5.33 \\
\midrule
\textbf{ReBind-Edit (Ours)} & \textbf{5.61} & \textbf{7.33} & \textbf{6.72} & \textbf{8.11} & \textbf{6.76} & \textbf{7.45} & \textbf{5.78} & \textbf{7.34} & \textbf{6.89} & \textbf{5.66} & \textbf{5.93} & \textbf{6.12} & 7.71 & \textbf{6.59} & \textbf{6.95} & \textbf{5.36} & \textbf{6.97} & \textbf{6.41} \\
\bottomrule
\end{tabular}
}
\vspace{-0.1in}
\end{table*}
\begin{table}[t]
\centering
\caption{Comparison of different MLLMs on dense video editing instruction generation. Evaluated on annotated UniVBench with GPT-5.1 as the judge. \textbf{Bold}: best.}
\label{tab:understanding_model_comparison}
\resizebox{\columnwidth}{!}{
\renewcommand{\arraystretch}{1.25}
\begin{tabular}{l|c|ccc}
\toprule
\textbf{Model} & \textbf{Size} & \textbf{IQ}$\uparrow$ & \textbf{RA}$\uparrow$ & \textbf{Overall}$\uparrow$ \\
\midrule
\multicolumn{5}{l}{\textit{Closed-Source Models}} \\
\textcolor{gray}{Gemini-2.5-Pro~\cite{comanici2025gemini}} & \textcolor{gray}{-}&\textcolor{gray}{46.37} & \textcolor{gray}{63.85} &\textcolor{gray}{55.11} \\
\textcolor{gray}{Gemini-3.1-Pro~\cite{gemini31pro}} & - & \textcolor{gray}{59.24} & \textcolor{gray}{52.90}& \textcolor{gray}{56.07} \\
\midrule
\multicolumn{5}{l}{\textit{Open-Source Models}} \\
Qwen2.5-Omni-3B~\cite{Qwen2.5-Omni} & 3B & 10.10&21.86 & 15.98\\
Qwen2.5-Omni-7B~\cite{Qwen2.5-Omni} & 7B & 14.32 &35.37 & 24.84 \\
Qwen3-Omni-Captioner~\cite{xu2025qwen3} & 30B-A3B & 22.01 & 39.47 & 30.74 \\
Qwen3-Omni-Instruct~\cite{xu2025qwen3} & 30B-A3B & 31.82 & 49.59 & 40.70 \\
Gemma4-31B-it~\cite{gemma4} & 31B & 49.49 & 36.71 & 43.10 \\
\midrule
\textbf{ReBind-Instruct (Ours)} & 30B-A3B & \textbf{61.68} & \textbf{59.60} & \textbf{61.18}\\
\bottomrule
\end{tabular}
}
\vspace{-0.1in}
\end{table}
\begin{table}[t]
\centering
\caption{Impact of instruction design on editing quality in multi-reference image-conditioned video editing tasks. Results are evaluated on the UniVBench benchmark.}
\label{tab:instruction_ablation}
\resizebox{\columnwidth}{!}{
\renewcommand{\arraystretch}{1.25}
\begin{tabular}{l|ccccccccc}
\toprule
\textbf{Instruction Type} &\textbf{Sub.} & \textbf{Back.} & \textbf{Action} & \textbf{Cam.} & \textbf{Color} & \textbf{Light.} & \textbf{Style} & \textbf{Rel.Pos.} & \textbf{Overall} \\
\midrule
Basic (Qwen3-Omni-Instruct) & 3.85 & 4.08 & 4.92 & 6.95 & 4.96 & 5.62 & 3.94 & 5.82 & 5.02 \\
\midrule
Dense (Qwen3-Omni-Instruct) & 4.36 & 4.41 & 5.38 & 7.21 & 5.51 & 6.18 & 4.44 & 6.27 & 5.47 \\
Dense (ReBind-Edit) &\textbf{5.08} & \textbf{5.38} & \textbf{5.76} & \textbf{7.46} & \textbf{6.07} & \textbf{6.57} & \textbf{5.06} & \textbf{6.67} & \textbf{5.88} \\
\bottomrule
\end{tabular}
}
\vspace{-2mm}
\end{table}
\subsection{Implementation Details}
\noindent \textbf{ReBind-Instruct.}
We build ReBind-Instruct upon Qwen3-Omni-30B-A3B-Instruct, a pre-trained vision-language model with audio-visual understanding capabilities. We fully fine-tune the model throughout both training stages. Training proceeds through a two-stage progressive scheme: Stage 1 with supervised fine-tuning on 40K video editing triplets for 3 epochs at batch size 128 and learning rate 1e-5, followed by Stage 2 with GRPO on 2K samples with 8 candidates per sample at learning rate 5e-6 and KL penalty $\beta=0.04$ on 32 H200 GPUs.

\noindent \textbf{ReBind-Edit.}
We build ReBind-Edit on top of LTX-2.3~\cite{hacohen2026ltx}, using Gemma-3-12B-IT~\cite{sellergren2025medgemma} as the text encoder and its 14B-parameter DiT backbone, which is fully fine-tuned. We train on approximately 100k video editing triplets with a batch size of 8 and a learning rate of 1e-6. For all editing tasks, we generate videos at 720P resolution with 81 frames, following the LTX-2.3 configuration.

\subsection{Evaluation Setup}

\noindent \textbf{Benchmarks.}
We evaluate on IntelligentVBench with single-reference cases covering object addition, removal, replacement, and UniVBench with single-reference and multi-reference cases across eight dimensions. Since UniVBench lacks dense instructions with explicit reference attribution, we manually annotate 1,240 samples in total covering both single-reference and multi-reference scenarios, with human annotators analyzing video pairs and reference images to generate structured instructions with explicit reference attribution, followed by expert review to ensure annotation quality and consistency.

\noindent \textbf{Metrics.}
For evaluating instruction quality, we propose two metrics.
Instruction Quality measures content completeness: we use GPT-5.1 to compare the generated instruction against ground-truth keypoints, identifying missing content, incorrect descriptions, and hallucinated elements, with the score reflecting the ratio of correctly captured keypoints.
Reference Accuracy measures reference attribution accuracy through structured triple matching: we extract object-descriptor-anchor triples from both generated and ground-truth instructions, then compute the F1 score based on exact matches where all three components must agree.
For video editing quality, we follow IntelligentVBench to evaluate Instruction Following, Condition Preserving, and Visual Quality, and follow UniVBench to evaluate across eight dimensions: Subject, Background, Action, Camera, Color, Lighting, Style, and Relative Position, with GPT-5.1 as the judger.

\noindent \textbf{Baselines.}
For reference-grounded instruction generation, we compare with general-purpose vision-language models Gemini-2.5-Pro~\cite{comanici2025gemini}, Gemini-3.1-Pro, Qwen2.5-Omni-3B, Qwen2.5-Omni-7B, Qwen3-Omni-Captioner-30B-A3B, Qwen3-Omni-Instruct-30B-A3B, and Gemma4-31B-it~\cite{sellergren2025medgemma}. For video editing on IntelligentVBench and UniVBench, we compare with open-source methods UniVideo~\cite{wei2025univideo}, VINO~\cite{chen2026vino}, Kiwi-Edit~\cite{lin2026kiwi}, and OmniWeaving~\cite{pan2026omniweaving}, along with closed-source models HappyHorse 1.0~\cite{happyhorse} and Kling-o3~\cite{klingo3}.

\subsection{Reference-Image Conditioned Video Editing Results}
Table~\ref{tab:editing_results} presents quantitative results on IntelligentVBench. Our method achieves the best overall average performance among all methods including closed-source systems. Specifically, we outperform the closed-source Kling-o3 model by 0.17 in overall average, and achieve 0.64 improvement over HappyHorse 1.0. Compared to the strongest open-source baseline OmniWeaving, we demonstrate a 0.49 improvement. These results establish that our dense instruction approach with explicit reference attribution is highly effective for video editing. Qualitative results in the Appendix further demonstrate visually coherent edits with accurate attribute extraction from reference images.
Table~\ref{tab:multref_comparison} presents results on UniVBench for both reference video editing (RVE) and multi-reference video editing (MRVE). Our method achieves the best overall performance among open-source methods, with 0.41 improvement over the strongest baseline UniVideo (hidden) on RVE and 0.60 on MRVE. These results validate that dense instructions with explicit reference attribution substantially improve editing quality, particularly in multi-reference scenarios where precise coordination across visual sources is critical.

Qualitative results reveal systematic failure modes across baselines: OmniWeaving exhibits cross-reference confusion, Kiwi-Edit shows minimal editing with poor reference consistency, and UniVideo suffers from severe temporal inconsistencies. Even closed-source systems struggle with identity preservation across frames. In contrast, our method produces accurate multi-reference coordination with correct feature attribution and strong temporal consistency, validating the effectiveness of dense instructions with explicit reference tokens.
\begin{table}[t]
\centering
\caption{Ablation study on MLLM training stages and reward components of ReBind-Instruct.}
\label{tab:vlm_training_stages}
\resizebox{\columnwidth}{!}{
\renewcommand{\arraystretch}{1.25}
\begin{tabular}{l|ccc|ccc}
\toprule
\multirow{2}{*}{\textbf{Model}} & \multicolumn{3}{c|}{\textbf{Reward}} & \multicolumn{3}{c}{\textbf{Metrics}} \\
\cmidrule(lr){2-4} \cmidrule(lr){5-7}
 & $\mathcal{R}_{D}$ & $\mathcal{R}_{C}$ & $\mathcal{R}_{L}$ & \textbf{IQ}$\uparrow$ & \textbf{RA}$\uparrow$ & \textbf{Overall}$\uparrow$ \\
\midrule
Qwen3-Omni-Instruct & -- & -- & -- & 31.82 & 49.59 & 40.70 \\
\midrule
Ours-Stage1 & -- & -- & -- & 56.84 & 53.11 & 54.98 \\
\midrule
\multirow{3}{*}{Ours-Stage2} & \checkmark & -- & -- & 63.81 & 54.15 & 58.98 \\
 & \checkmark & \checkmark & -- & 60.74 & \textbf{60.07} & 60.40 \\
 & \checkmark & \checkmark & \checkmark & \textbf{61.68} & 59.60 & \textbf{61.18} \\
\bottomrule
\end{tabular}
}
\vspace{-0.1in}
\end{table}
\subsection{Dense Instruction Generation Quality}
Table~\ref{tab:understanding_model_comparison} evaluates our dense instructor against state-of-the-art vision-language models on the annotated UniVBench multi-reference subset. Our method achieves state-of-the-art performance among all models, including closed-source systems. Compared to the strongest open-source baseline Gemma4-31B-it, our method achieves 18.8\% overall improvement. Compared to the closed-source Gemini-3.1-Pro, we demonstrate 5.11\% overall improvement. This validates our core hypothesis that generating reference-grounded editing instructions with explicit visual anchoring requires specialized training beyond general-purpose MLLM pretraining.
\subsection{Ablation Study}
\subsubsection{Impact of Dense Instructions on Editing Quality}
To isolate the impact of instruction design and quality on editing performance, we train ReBind-Edit on 50k randomly sampled training cases for 5k steps with three types of instructions: basic instructions from Qwen3-Omni-Instruct, dense instructions with explicit reference tokens from the same Qwen3-Omni-Instruct, and dense instructions from our ReBind-Instruct.

Table~\ref{tab:instruction_ablation} shows that instruction structure matters: dense instructions improve editing quality by 0.45 over basic instructions using the same MLLM. Moreover, instruction quality matters: our ReBind-Instruct further improves results by 0.41 over Qwen3-Omni-Instruct with the same dense format. These results demonstrate that higher-quality instructions directly translate to better editing performance.
\begin{figure}[t]
\centering
\includegraphics[width=\columnwidth]{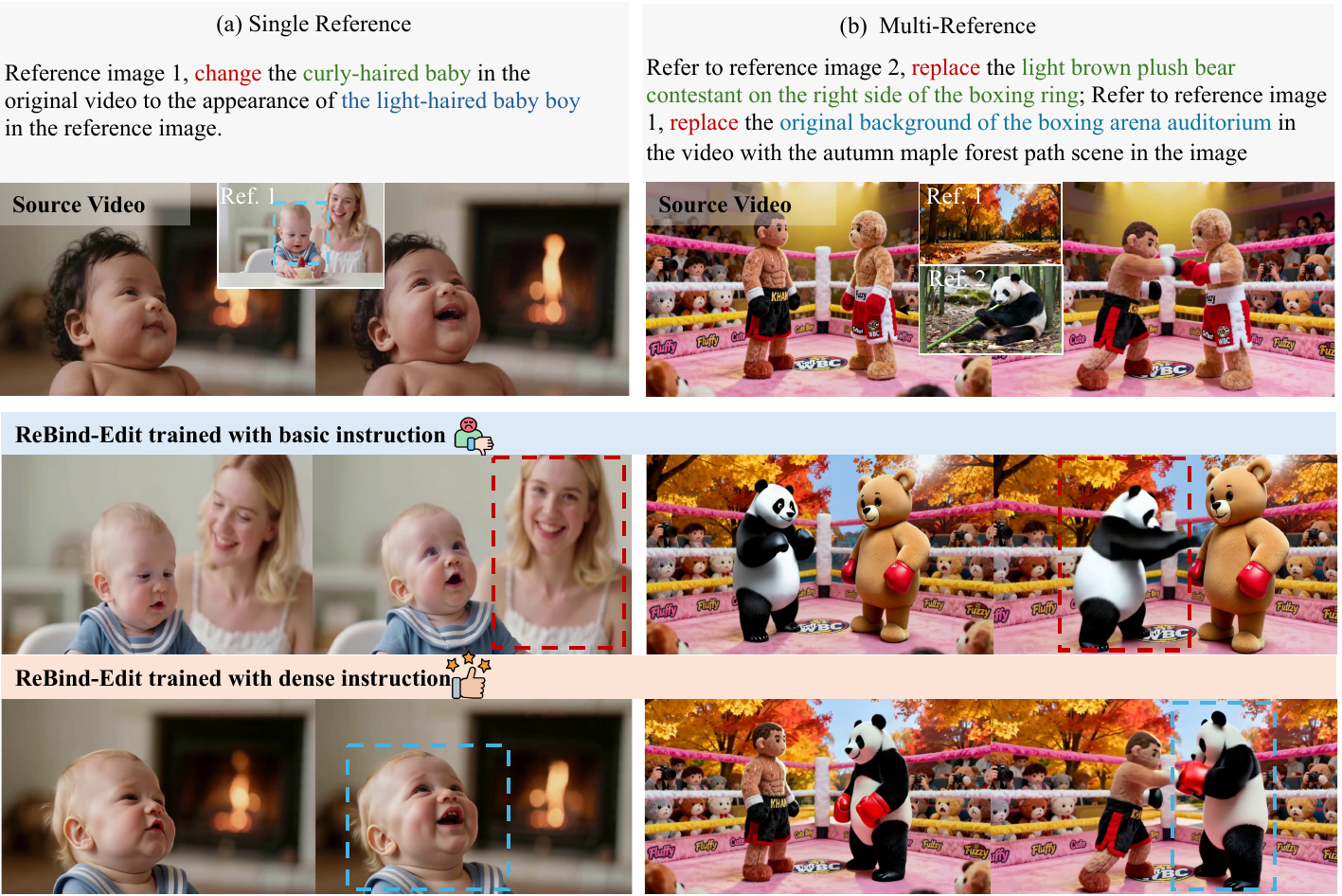}
\caption{Qualitative comparison of our ReBind-Edit trained with different instruction types.}
\label{fig:instruction_ablation_vis}
\vspace{-3mm}
\end{figure}
\subsubsection{Impact of Different Reward Components}
In Table~\ref{tab:vlm_training_stages}, we conduct an in-depth analysis of each component within our GRPO training pipeline. Stage 1 supervised fine-tuning yields substantial improvements over the base Qwen3-Omni-Instruct model with 14.28\% overall gain, demonstrating the effectiveness of learning structured instructions on curated editing triplets. During Stage 2 GRPO optimization, we progressively add reward components. The dual checklist rewards $\mathcal{R}_{D}$ contribute 4.00\% overall gain, the referential consistency reward $\mathcal{R}_{C}$ adds 1.42\%, and the length reward $\mathcal{R}_{L}$ contributes 0.78\%. These results validate that multi-objective optimization through tailored rewards is essential for reference-grounded instruction generation.

\subsection{Conclusion}
We address multi-reference video editing by introducing dense instructions with explicit reference attribution as the intermediate representation. Our key insight is that the bottleneck lies in generating appropriate instructions rather than generation model capacity. We develop a two-stage training pipeline to generate instructions with explicit visual reference tokens completely and accurately. We further design a editor model that binds visual features to textual anchors through cross-attention mechanisms. Experiments demonstrate state-of-the-art performance on both instruction generation and video editing benchmarks, validating that explicit reference grounding is essential for controllable multi-reference video editing.

    \small
    \bibliographystyle{ieeenat_fullname}
    \bibliography{main}
\clearpage
\appendix
\section*{Appendix}
\addcontentsline{toc}{section}{Appendix}

\section{Evaluation Benchmark Details}

To complement the evaluation setup described in Section 5.2, we provide detailed information about the benchmarks used in our experiments.

\subsection{IntelligentVBench}
IntelligentVBench~\cite{pan2026omniweaving} is a comprehensive benchmark designed to assess next-level intelligent unified video generation by integrating abstract reasoning and free-form composition. We evaluate our method on the Text-Image-Video-to-Video (TIV2V) subset, which assesses cross-modal compositional editing capabilities. The TIV2V task requires models to extract specific concepts from reference images and integrate them into source videos while preserving temporal consistency and visual fidelity of non-target elements. This subset comprises 210 test cases spanning three operational objectives: inserting an object from a reference image into the source video (71 cases), substituting a specific visual element within the video with one from the reference image (73 cases), and replacing the video's original background with the scene depicted in the reference image (66 cases). The benchmark employs Gemini-2.5-Pro as an automatic evaluator with three metrics rated on a 1-5 scale: Instruction Following (IF) measures the model's capacity to seamlessly integrate video and image in strict adherence to instructions, Condition Preserving (CP) evaluates preservation of the source video's unedited regions and verification that newly introduced elements faithfully conform to reference images, and Visual Quality (VQ) assesses aesthetic quality, temporal consistency, and motion naturalness. Overall performance is measured by AVG = (IF + CP + VQ)/3 and MIN = Min(IF, CP, VQ).

\subsection{UniVBench}
UniVBench~\cite{wei2026univbench} is the first unified benchmark providing comprehensive evaluation across video understanding, generation, and editing tasks with multi-shot support and copyright-free content. For reference-guided video editing evaluation, UniVBench employs an agentic evaluation system UniV-Eval that decomposes videos into shot-level units and performs fine-grained assessment across nine major category groups: Subject, Relative Position, Actions, Background, Color, Lighting, Style, Atmosphere, and Camera. These categories are further decomposed into 21 specific sub-dimensions that enable diagnostic feedback at the shot level. The evaluation agent performs per-category comparisons between model outputs and input tuples (source video, reference images, text instructions), producing structured weakness checklists that highlight fine-grained deficiencies. Final scores are aggregated along multiple evaluation dimensions, with each dimension scored on a scale from 0 to 10. While UniVBench originally uses Seed-1.6 as the evaluation LLM, we adopt GPT-5.1 for our experiments to ensure more reliable and consistent evaluation quality. UniVBench contains diverse test cases covering both single-reference and multi-reference video editing scenarios. For our evaluation, we manually annotate 1,240 samples (620 single-reference and 620 multi-reference) with dense instructions featuring explicit reference attribution to enable fair comparison of instruction generation quality and video editing performance.

\section{Baseline Methods Details}

To provide context for the comparative evaluation in Section 5, we present detailed descriptions of all baseline methods compared in our experiments.

\textbf{UniVideo}~\cite{wei2025univideo} is a unified framework that integrates video understanding, generation, and editing within a single architecture. It adopts a dual-stream design combining a Multimodal Large Language Model (MLLM) for instruction understanding with a Multimodal DiT (MMDiT) for video generation. UniVideo supports two distinct conditioning modes for reference image integration: (1) \textit{query-based} mode extracts reference features through learned query tokens and injects them via cross-attention, enabling flexible composition, and (2) \textit{hidden-state-based} mode directly incorporates the MLLM's hidden states into the generation process, providing richer semantic guidance. These two modes offer different trade-offs between compositional flexibility and semantic fidelity.

\textbf{Kiwi-Edit}~\cite{lin2026kiwi} is a unified video editing framework designed for instruction-reference-following tasks with single reference images. The method synergizes learnable queries and latent visual features for reference semantic guidance, processing reference images through dual pathways: learnable query tokens for flexible composition and latent visual features for precise visual control. Features are fused with the video generation backbone through cross-attention mechanisms. While originally designed for single-reference scenarios, we evaluate its performance on multi-reference editing tasks to assess its generalization capability.

\textbf{OmniWeaving}~\cite{pan2026omniweaving} is a unified video generation model that combines a multimodal large language model with a latent diffusion model for video synthesis. The framework integrates video generation and editing tasks within a single architecture, trained on massive-scale pretraining data encompassing compositional and reasoning-augmented scenarios. The model employs attention-based feature fusion mechanisms to process interleaved text, multi-image, and video inputs. We evaluate its generalization capability on multi-reference video editing tasks.

\textbf{VACE}~\cite{jiang2025vace} is an all-in-one framework for video creation and editing built on Diffusion Transformers. VACE unifies multiple video synthesis tasks including reference-to-video generation, video-to-video editing, and masked video-to-video editing through a Video Condition Unit (VCU) that organizes diverse task inputs into a unified interface. The framework employs a Context Adapter structure to inject task-specific concepts via formalized temporal and spatial representations. We evaluate two variants: VACE-Wan2.1 built on Wan2.1 and VACE-LTX built on LTX.

\textbf{VINO}~\cite{chen2026vino} is a unified visual generator that performs image and video generation and editing within a single framework. VINO couples a vision-language model with a Multimodal Diffusion Transformer (MMDiT), where multimodal inputs (text, images, videos) are encoded as interleaved conditioning tokens to guide the diffusion process. This design supports multi-reference grounding, long-form instruction following, and coherent identity preservation across static and dynamic content without modality-specific architectural components.
\section{Additional Implementation Details}

To ensure reproducibility of the results presented in Section 5, we provide comprehensive training hyperparameters for all three training stages described in Section 4. This section details the optimization settings, learning rate schedules, parallelization strategies, and GRPO-specific parameters used in our experiments.

Table~\ref{tab:training_hyperparameters} presents the detailed training hyperparameters across all three training stages. Stage 1 uses supervised fine-tuning on ReBind-Instruct with a learning rate of $1.0 \times 10^{-5}$ and cosine learning rate scheduling with warmup for 3 epochs. Stage 2 applies GRPO reinforcement learning to further optimize ReBind-Instruct with GRPO-specific parameters such as beta=0.04 for KL penalty control and 8 generations per training step. Stage 3 trains ReBind-Edit using dense instructions from ReBind-Instruct with the same base learning rate of $1.0 \times 10^{-5}$ but with a warmup-constant-cosine scheduler for 15K steps. All stages use mixed precision training and employ gradient checkpointing for memory efficiency.

\begin{table}[h]
\centering
\caption{Training hyperparameters across three stages: Stage 1 (ReBind-Instruct SFT), Stage 2 (ReBind-Instruct GRPO), and Stage 3 (ReBind-Edit).}
\label{tab:training_hyperparameters}
\resizebox{\columnwidth}{!}{
\begin{tabular}{lccc}
\toprule
\textbf{Hyperparameters} & \textbf{Stage 1} & \textbf{Stage 2} & \textbf{Stage 3} \\
 & \textbf{(Instructor SFT)} & \textbf{(Instructor GRPO)} & \textbf{(Editor)} \\
\midrule
Learning rate & $1.0 \times 10^{-5}$ & $1.0 \times 10^{-5}$ & $1.0 \times 10^{-5}$ \\
LR scheduler & Cosine & Cosine & Warmup-Constant-Cosine \\
LR warmup steps & 0.02 (fraction) & - & 500 \\
Constant steps & - & - & 500 \\
Min learning rate & $4.0 \times 10^{-6}$ & - & $1.0 \times 10^{-6}$ \\
Weight decay & 0.0 & 0.0 & - \\
Optimizer & AdamW & AdamW & AdamW \\
Global batch size & 128 & 128 & 8 \\
Micro batch size & 4 & 1 & 1 \\
Gradient accumulation & - & - & 8 \\
Max sequence length & 32768 & 32768 & - \\
MOE aux loss coeff & $1.0 \times 10^{-3}$ & $1.0 \times 10^{-3}$ & - \\
\midrule
\multicolumn{4}{l}{\textit{Parallelization (Stages 1-2 only)}} \\
\midrule
Context parallel size & 1 & 1 & - \\
Tensor parallel size & 4 & 4 & - \\
Pipeline parallel size & 1 & 1 & - \\
Expert parallel size & 8 & 8 & - \\
\midrule
\multicolumn{4}{l}{\textit{GRPO-specific parameters (Stage 2 only)}} \\
\midrule
Beta & - & 0.04 & - \\
Epsilon (low, high) & - & 0.005, 0.01 & - \\
Num generations & - & 8 & - \\
Temperature & - & 1.0 & - \\
Importance sampling & - & sequence & - \\
\bottomrule
\end{tabular}
}
\end{table}

\section{Details of Training Dataset}

In this section, we present detailed specifications of our data collection, filtering, and annotation procedures.
\subsection{Target Video Selection Criteria}

Target videos are collected from OpenVid-1M and AVSET-10M based on content suitability for editing operations. We prioritize videos with clear subjects and stable camera motion to facilitate reliable reverse editing. Specifically, videos are filtered to ensure: (1) minimum resolution of 720P and duration between 2-10 seconds to balance visual quality with computational efficiency, (2) presence of identifiable entities (people, objects, or scenes) suitable for attribute-level editing, and (3) absence of extreme motion blur or occlusion that could compromise editing quality. Videos containing predominantly text overlays or with severe compression artifacts are excluded through automated quality checks.

\subsection{Inverse Instruction Design}

To generate source videos through reverse editing, we design inverse instructions that specify attribute transformations opposite to the desired editing direction. For each target video, we first identify editable attributes using automated captioning models, then construct inverse instructions following templates such as "change the [entity] to have [inverse attribute]" or "replace the [entity] with one having [contrasting property]." For example, if the target video shows a person wearing red clothing, the inverse instruction might specify changing to blue clothing. This ensures the generated source video provides a meaningful starting point for forward editing tasks.

\subsection{Quality Filtering Thresholds}

In the video-to-video similarity filtering stage, we compute cosine similarity between CLIP~\cite{radford2021learning} embeddings extracted from source and target videos. Pairs with similarity below 0.3 indicate failed edits where the source and target are semantically unrelated, while similarity above 0.95 suggests insufficient editing variation. Both extremes are filtered to ensure the dataset contains meaningful editing challenges. For MLLM verification, GPT-5.1 is prompted to evaluate whether the visual changes between source and target align with the inverse editing instruction. Samples where the model identifies semantic inconsistencies, incomplete edits, or visual artifacts are flagged and excluded. This two-stage process removes approximately 15\% of initial pairs.

\subsection{Reference Image Selection Criteria}

From the multi-frame mask candidates generated by SAM3 and inpainted by Flux2, we select final reference images based on visual quality metrics. Specifically, we prioritize: (1) front-facing views where the entity is clearly visible with minimal occlusion (occlusion ratio below 20\%), (2) high visual clarity measured by Laplacian variance to exclude blurry frames, and (3) completeness where the entity is fully within the frame boundary. For each entity, we select the highest-quality image as reference. In multi-reference scenarios involving 2-5 reference images per sample.

\subsection{Dataset Statistics}

The final dataset contains approximately 100K video editing triplets with the following distribution: 52.3\% single-reference samples (one entity edited) and 47.7\% multi-reference samples (2-3 entities edited simultaneously). Videos range from 2-10 seconds in duration with resolutions between 480p and 1080p. Content categories include people (49.2\%), animals (12.6\%), objects (32.8\%), and scenes (5.4\%). Each sample is associated with 1-5 reference images and corresponding dense instruction annotations generated through the two-stage training pipeline described in Section 3.2.

\subsection{Human Annotation for GRPO Training} To provide high-quality supervision for Stage 2 GRPO training, we recruit eight experienced annotators to manually create dense instructions for 2K samples from the synthetic dataset. Annotators are provided with the source video, target video, and reference images, and are tasked with writing dense instructions that: (1) explicitly bind visual attributes to reference images using standardized tokens (e.g., \texttt{<image\_1>}, \texttt{<image\_2>}), (2) clearly distinguish between preserved and modified elements, and (3) provide sufficient detail to guide the editing operation. While these ground-truth instructions are human-annotated, auxiliary training signals such as checklists and reward scores are automatically generated by GPT-5.1 for scalability.

\subsection{Human Annotation for UniVBench Evaluation} To evaluate instruction generation performance on UniVBench, we manually annotate 1,240 samples from the benchmark dataset. While UniVBench provides basic instructions specifying high-level editing intents, these lack explicit reference attribution and detailed descriptions. Our annotators directly observe the source videos, target videos, and reference images to write new dense instructions from scratch, ensuring: (1) explicit reference tokens bind each visual attribute to its corresponding reference image, (2) the preserve-vs-change distinction is clearly articulated, and (3) instruction density and format match our training data. This process generates gold-standard dense instructions for UniVBench, enabling fair evaluation of our model's instruction generation capabilities.

\section{Training Curve of ReBind-Edit}
To validate the training stability and convergence of ReBind-Edit reported in Section 5.4.1, we present the training loss curve on the 50k subset used for ablation experiments. This curve demonstrates that the model reaches stable convergence by 5000 steps, justifying our choice of this checkpoint for fair comparison across different instruction types.

\begin{figure}[h]
\centering
\includegraphics[width=\columnwidth]{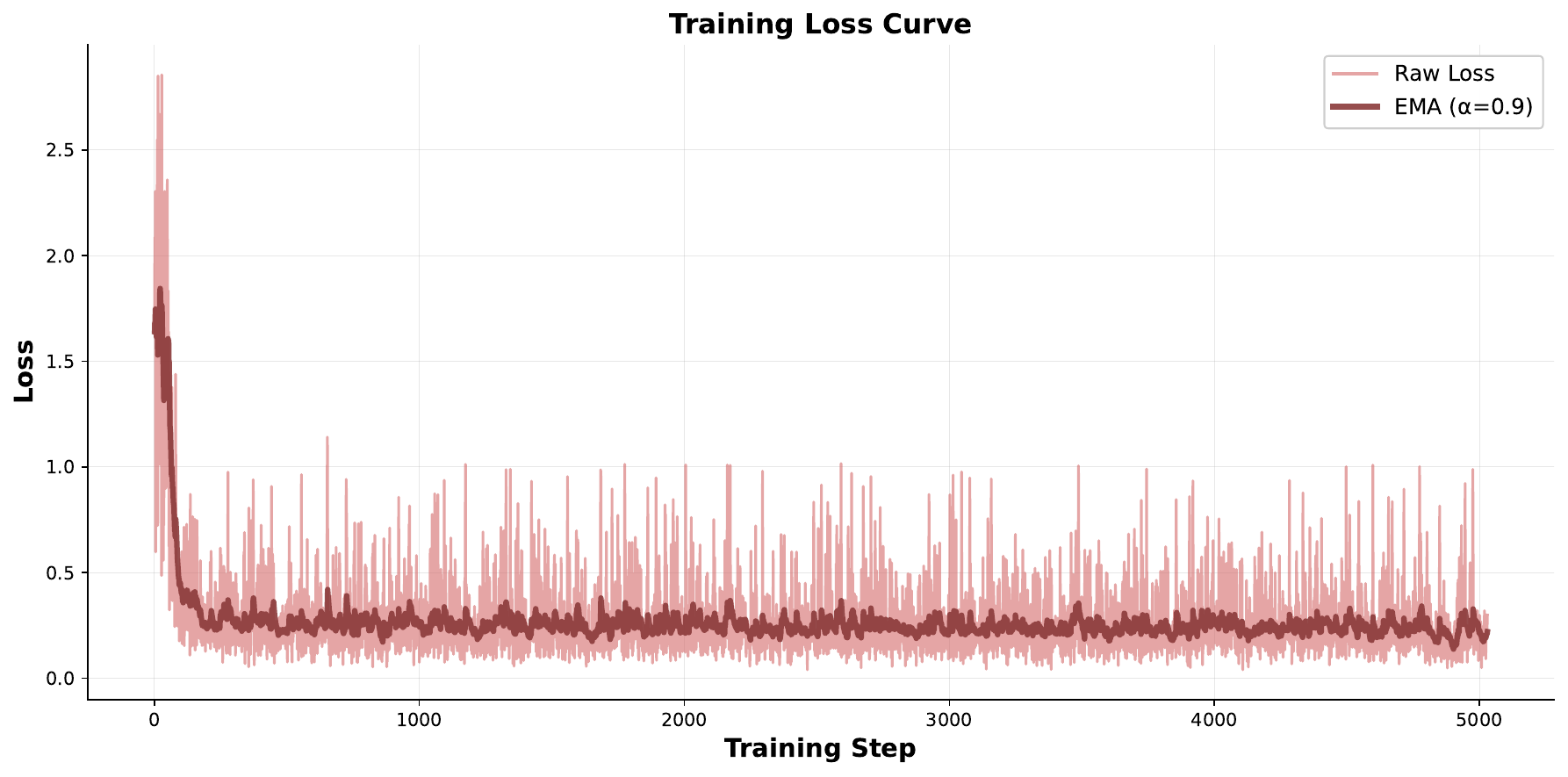}
\caption{Training loss curve of ReBind-Edit with dense instructions and explicit reference tokens.}
\label{fig:training_loss}
\vspace{-3mm}
\end{figure}

Figure~\ref{fig:training_loss} shows the training loss curve of ReBind-Edit when trained on 50k randomly sampled training cases with dense instructions and explicit reference tokens. The loss converges rapidly in early training and reaches a stable plateau after approximately 3000 steps, with minimal fluctuation through 5000 steps. This convergence pattern validates our choice of using 5000-step checkpoints for the ablation experiments of instruction type, ensuring all instruction design variants are evaluated at a well-converged and stable training state for fair comparison.

\section{Reward Curves of ReBind-Instruct}

To provide insights into the GRPO training dynamics of ReBind-Instruct described in Section 4.2, we present the reward curves across all training iterations. These curves demonstrate the effectiveness of our reward design, showing consistent improvement in dual checklist rewards, stable referential consistency, appropriate length control, and decreasing load balancing loss.

\begin{figure}[h]
\centering
\includegraphics[width=\columnwidth]{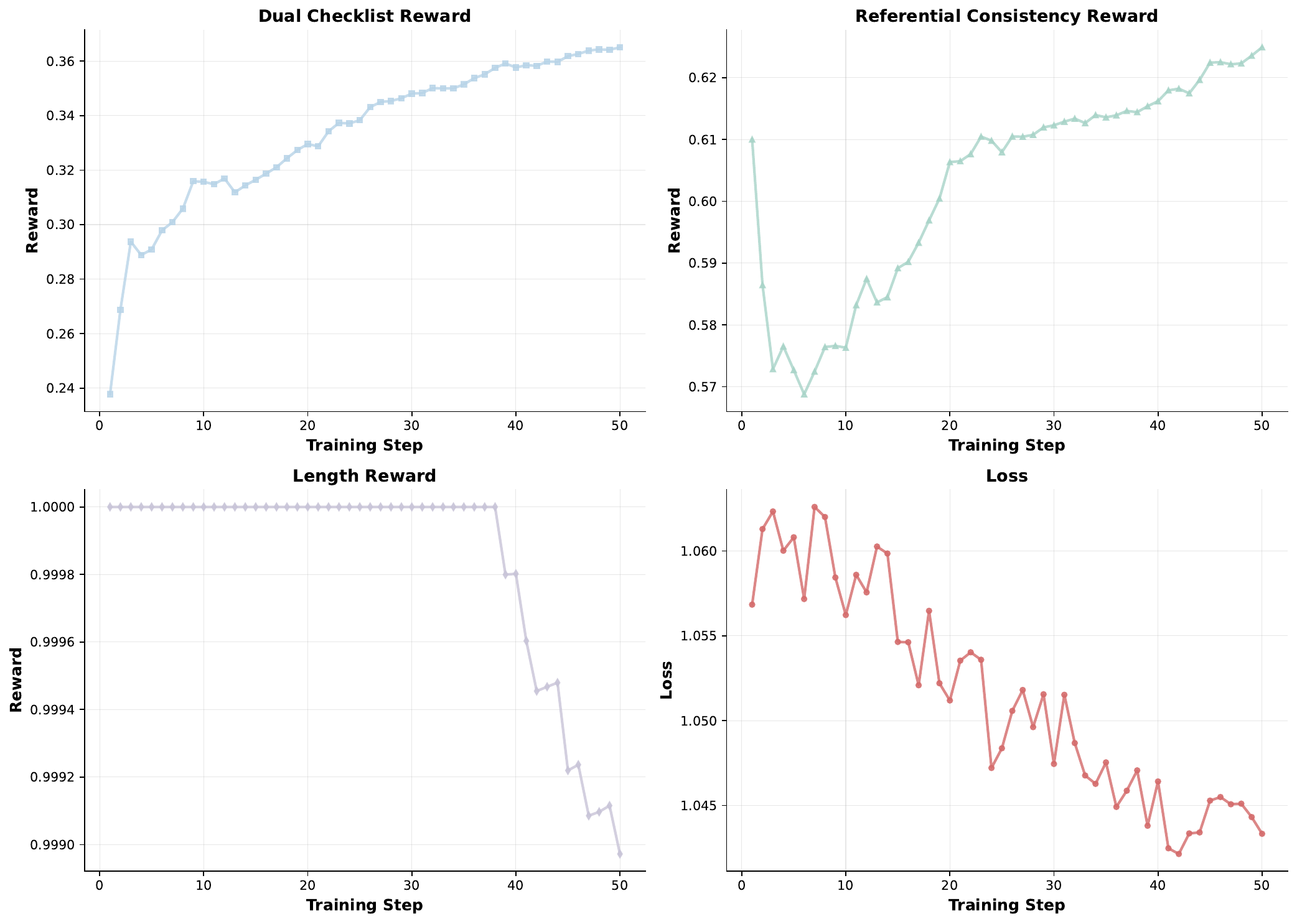}
\caption{Training dynamics of ReBind-Instruct.}
\label{fig:reward_curves}
\vspace{-3mm}
\end{figure}

Figure~\ref{fig:reward_curves} shows the training dynamics of ReBind-Instruct across four key metrics. The Dual Checklist Reward, which evaluates the balance between preserving unchanged content and accurately modifying specified elements, demonstrates consistent improvement throughout training, indicating effective learning of the preserve-vs-change distinction. The Referential Consistency Reward maintains high performance, validating that the model successfully learns coherent mappings between reference images and target attributes. The Length Reward remains consistently high, confirming that generated instructions meet the desired token length constraints. The Load Balancing Loss shows a gradual decrease, reflecting improved stability in the model's attention mechanisms during training.

\section{User Study}
To complement automated evaluation with human judgment, we conduct a user study with 15 participants on 50 randomly sampled cases from UniVBench. For each case, participants are shown the source video, reference images, editing instruction, and the anonymized outputs from all methods. They are asked to rank the results for each of the four evaluation criteria: Instruction Following, Source Video Preserving, Reference Image Preserving, and Visual Quality. We compare ReBind-Edit against three open-source baselines: UniVideo~\cite{wei2025univideo}, Kiwi-Edit~\cite{lin2026kiwi}, and OmniWeaving~\cite{pan2026omniweaving}. Preference scores are computed from the rankings, with higher scores indicating better performance.

\begin{figure}[h]
\centering
\includegraphics[width=\columnwidth]{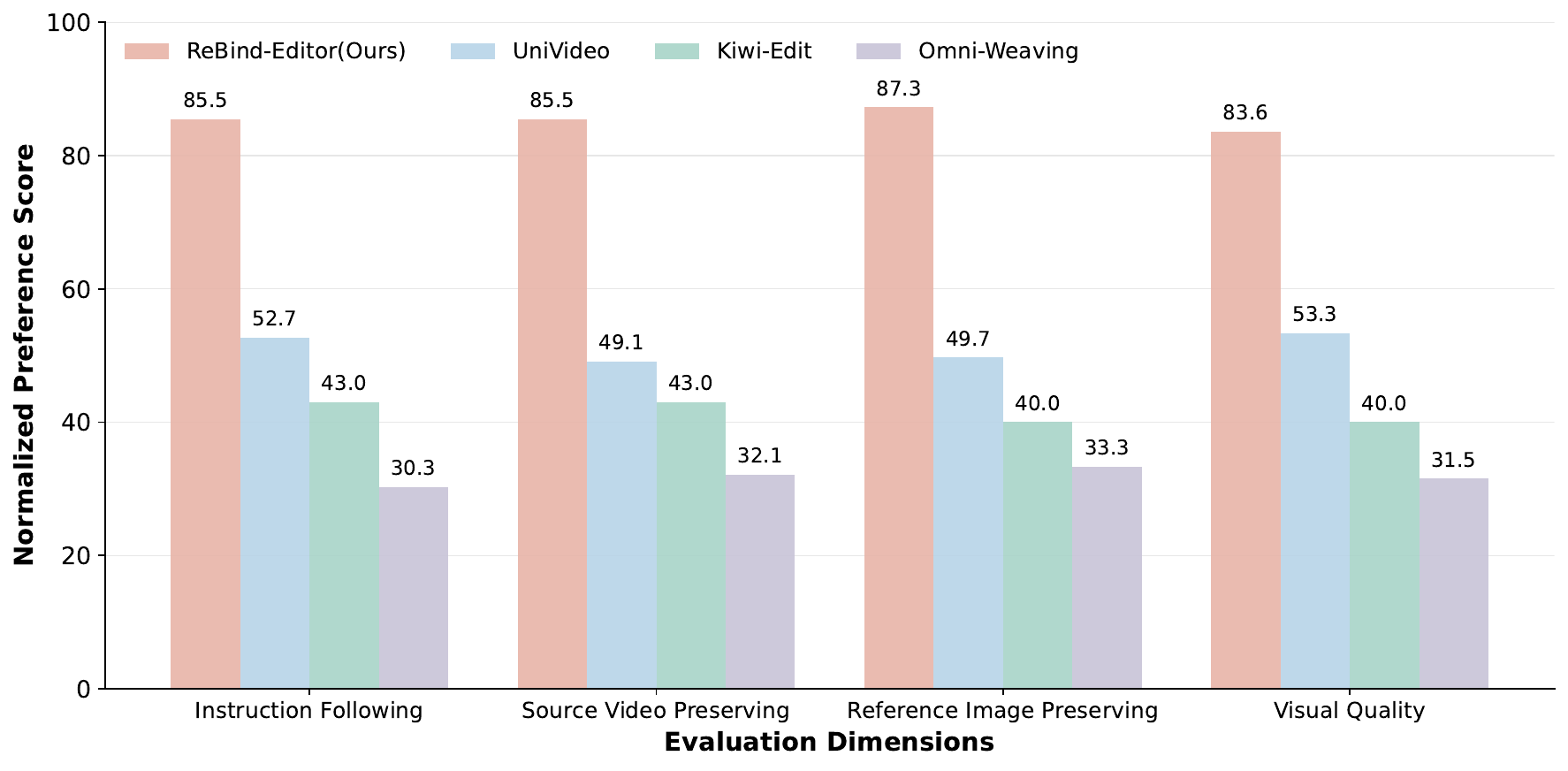}
\caption{User study results comparing ReBind-Edit against other representative models across four evaluation dimensions.}
\label{fig:user_study}
\vspace{-3mm}
\end{figure}

As shown in Figure~\ref{fig:user_study}, ReBind-Edit achieves the highest preference scores across all four evaluation dimensions. The advantage is most pronounced in Reference Image Preserving, validating the effectiveness of our dense instructions with explicit reference relationships for accurate multi-reference coordination. ReBind-Edit significantly outperforms all baselines, demonstrating superior performance in both instruction following and visual quality.

\section{More Qualitative Results}
To provide a more comprehensive view of ReBind's performance across diverse editing scenarios, we present additional qualitative comparisons on both single-reference image-conditioned video editing (RVE) and multi-reference image-conditioned video editing (MRVE) tasks. These examples further demonstrate the effectiveness of our dense instructions with explicit reference relationships in achieving precise visual attribute extraction and accurate multi-reference coordination.

\begin{figure*}[t]
\centering
\includegraphics[width=\textwidth]{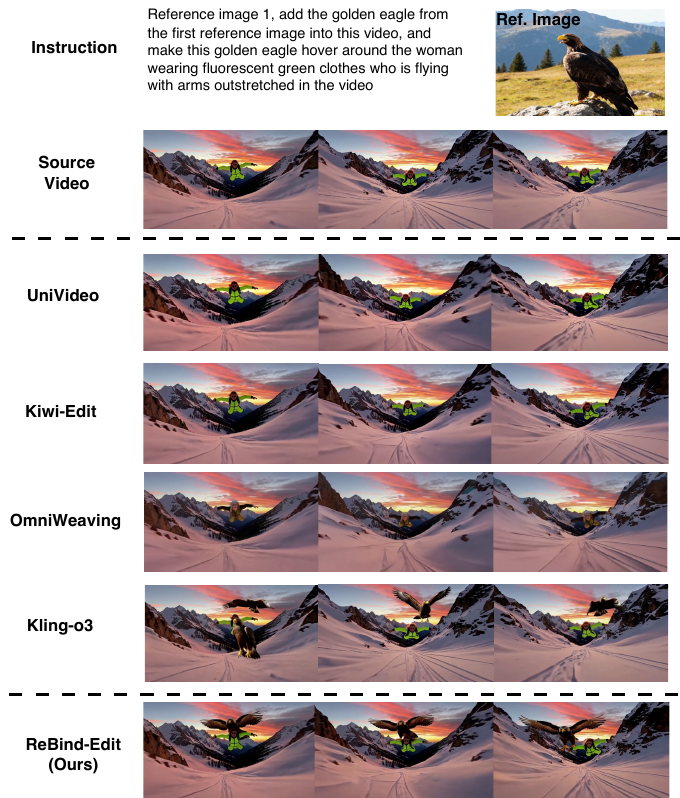}
\caption{Qualitative comparisons for single-reference image-conditioned video editing on UniVBench.}
\label{fig:single_ref_case}
\end{figure*}

\textbf{Single-Reference Editing.} Figure~\ref{fig:single_ref_case} presents additional comparisons on UniVBench for single-reference editing tasks, where we compare ReBind-Edit against UniVideo, Kiwi-Edit, OmniWeaving, and the closed-source Kling-o3. These examples reveal common challenges in reference-conditioned video editing: UniVideo and Kiwi-Edit sometimes struggle with complete instruction execution, where certain editing operations may not be fully realized in the output; OmniWeaving faces challenges in balancing editing execution with identity consistency across frames; Kling-o3 occasionally exhibits temporal instabilities in object generation during initial frames. In contrast, our method leverages explicit reference attribution in dense instructions to achieve more consistent editing results, with accurate attribute extraction from reference images, improved temporal coherence throughout the video sequence, and faithful execution of editing operations while preserving the original video's dynamics.

\begin{figure*}[t]
\centering
\includegraphics[width=\textwidth]{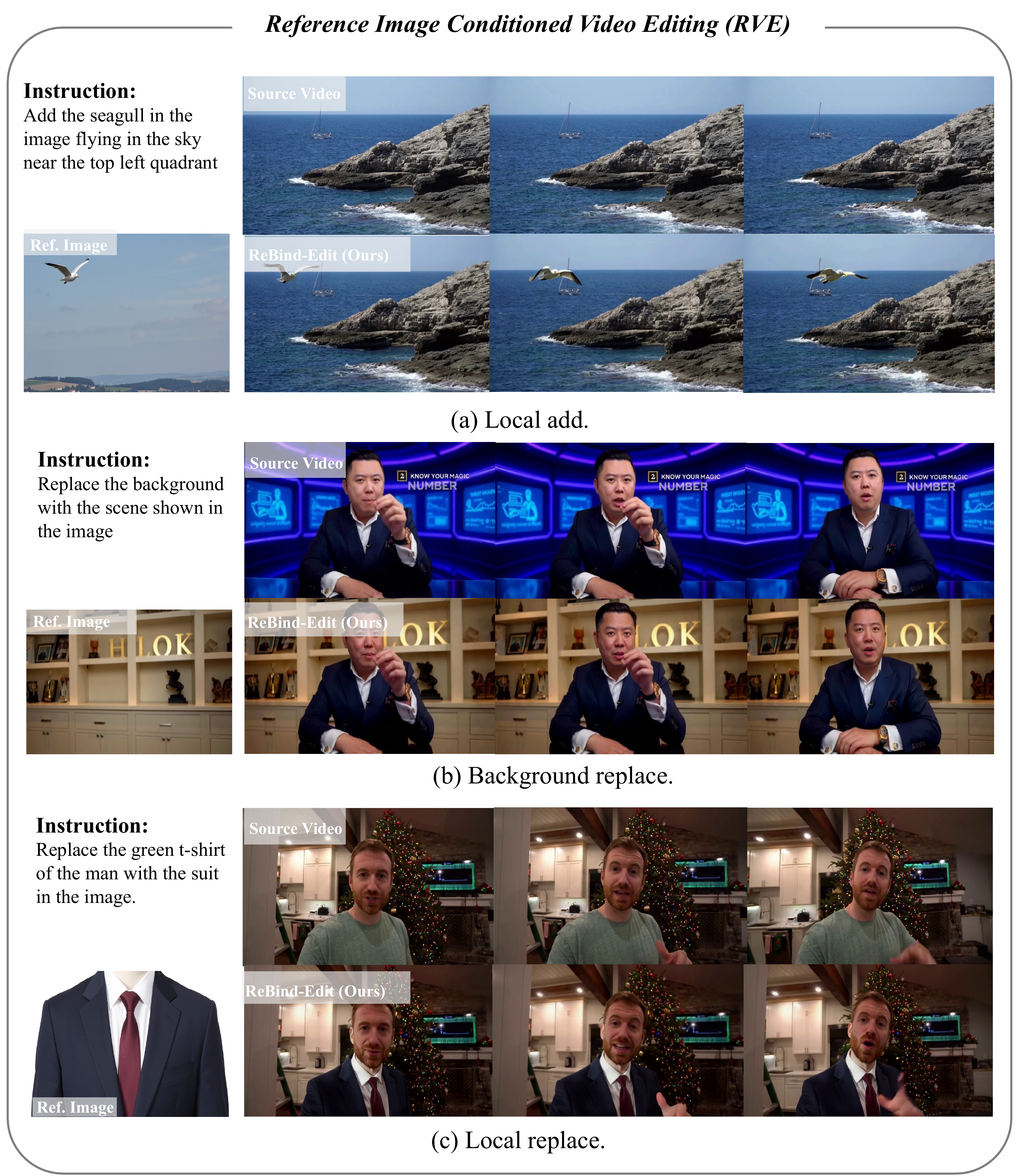}
\caption{Additional single-reference image-conditioned video editing (RVE) results on IntelligentVBench. Examples span diverse editing types including (a) local add, (b) background replacement, and (c) local replacement.}
\label{fig:single_ref_case4}
\end{figure*}

Figure~\ref{fig:single_ref_case4} showcases additional results on IntelligentVBench, covering diverse editing operations including local add, background replacement, and local replacement. These examples validate that our approach generalizes well across different editing types, consistently delivering visually coherent results that accurately follow both the editing instructions and reference visual characteristics.

\begin{figure*}[t]
\centering
\includegraphics[width=\textwidth]{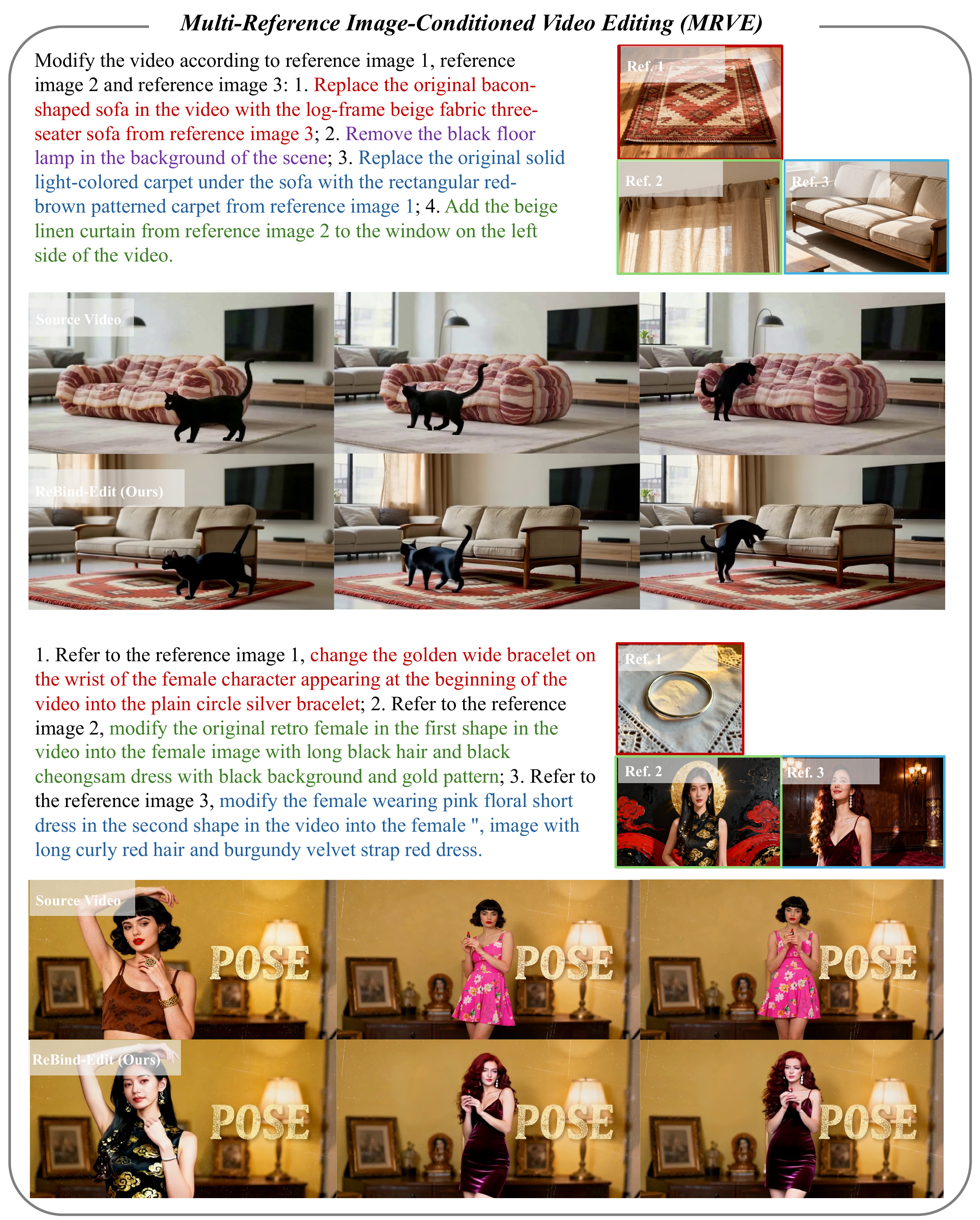}
\caption{Additional multi-reference image-conditioned video editing (MRVE) results on UniVBench.}
\label{fig:multi_ref_case_2}
\end{figure*}

\textbf{Multi-Reference Editing.} Figure~\ref{fig:multi_ref_case_2} provides supplementary multi-reference editing results from UniVBench. These cases involve coordinating visual attributes from multiple reference sources simultaneously, which presents significant challenges for cross-reference attribution. The results demonstrate that ReBind-Edit successfully handles complex multi-reference scenarios by correctly binding each visual attribute to its designated source through explicit reference tokens in dense instructions. Our method avoids the cross-reference confusion that commonly affects baseline approaches, producing edited videos where each modified element accurately reflects its corresponding reference image.

\onecolumn
\section{MLLM Prompts for Instruction Filtering}

To ensure the quality of generated dense instructions, we apply automated filtering through GPT-5.1 to detect and remove erroneous annotations. The filtering process checks for three categories of errors: reference token consistency, content quality, and format correctness.

\begin{tcolorbox}[breakable, enhanced jigsaw, colback=white!95!gray, colframe=gray!50!black, rounded corners, title={Instruction Quality Filtering Prompt}]
\footnotesize
You are a quality control specialist for video editing instructions. Your task is to evaluate whether a generated dense instruction meets quality standards for multi-reference video editing.

\textbf{Input:} Source Video Description, Available Reference Tokens (e.g., \texttt{<image\_1>}, \texttt{<image\_2>}, \texttt{<video\_1>}), Generated Instruction, Edit Type.

\textbf{Evaluation Criteria:}

\textbf{1. Reference Token Consistency:} All reference tokens must exist in the available list, follow correct format (\texttt{<image\_N>} or \texttt{<video\_N>}), and be correctly attributed. No hallucinated or missing tokens.

\textbf{2. Content Quality:} Length should be 50-300 tokens. No excessive repetition (same phrase more than 3 times). Logical flow and clear structure. Covers both preserved and changed elements.

\textbf{3. Format Correctness:} Proper grammar and sentence structure. Clear distinction between preserved and modified elements. No truncated or incomplete sentences.

\textbf{Decision Rules:} FILTER if: reference token errors, severe length anomalies (less than 20 or more than 512 tokens), high repetition, or critical format errors. KEEP if: all tokens are valid, content is coherent and complete.

Evaluate the given instruction and return whether it should be filtered with a brief explanation.
\end{tcolorbox}

\section{MLLM Prompts for UniVBench Evaluation}
For evaluating video editing quality on UniVBench, we employ GPT-5.1 as the judge model to assess edited videos across eight dimensions: Subject, Background, Action, Camera, Color, Lighting, Style, and Relative Position. The evaluation balances edit success with temporal consistency, recognizing that successful edit execution is the primary goal while maintaining appropriate consistency when the edit permits.
\begin{tcolorbox}[breakable, enhanced jigsaw, colback=white!95!gray, colframe=gray!50!black, rounded corners, title={Complete Multi-Reference Video Editing Evaluation Prompt}]
\footnotesize
You are a meticulous and detail-oriented bilingual (Chinese-English) "AI Film Quality Inspector."
Your sole mission is to evaluate the videos generated by a video understanding and generation model,
assessing whether they meet the final delivery standards in terms of technical execution, content fidelity, and artistic expressiveness.
You will receive an OriginalVideo (a list of timestamped sampled images), ReferenceImages, EditInstruction and a GeneratedVideo (also presented as timestamped sampled images).

---

Input Format Description

You will receive four core inputs:

1. Original Video (OriginalVideo)
   - A reference video provided by the user, given as sampled images with corresponding timestamps.
   - It serves as the semantic and informational baseline, defining what should be preserved or transformed.

2. Reference Images (ReferenceImages)
   - A set of images representing reference visual style, tone, frame composition, and subject references.
   - These images reflect the specific visual characteristics (e.g., subject appearance, lighting, color palette) the user hopes the generated video will achieve.

3. Edit Instruction (EditInstruction)
   - A natural language description indicating the edits or style adjustments the user wants to apply to the original video.
   - Example: "change daytime to nighttime", "replace the person with the subject from ReferenceImage1."
   - It may also work together with ReferenceImages for modification.

4. Generated Video (GeneratedVideo)
   - The output produced by the video understanding and generation model, also provided as timestamped sampled images.

---

Core Inspection Framework

Before checklist evaluation, internally decompose all input using the structured attributes below.

video\_attribute\_requirements (Static Video Attributes)
- subjects: quantity, gender, clothing, appearance, expressions, visible text/logos.
- background: time, location, architecture, objects, landscaping, indoor/outdoor elements.
- lighting: lighting\_direction, lighting\_effect, brightness, light source realism.
- color: overall\_tone, saturation, contrast, color harmony.
- image\_style: realistic, cinematic, anime, documentary, handheld aesthetic, etc.
- atmosphere: mood and emotional tone (e.g., warm, tense, nostalgic, mysterious).

relative\_position\_requirements (Spatial Semantics)
- inter\_frame\_layout: spatial continuity of subjects and environment.
- inter\_subject\_relation: body distance, facing directions, interpersonal dynamics.
- subject\_camera\_relation: subject-to-camera orientation and framing logic.

actions\_requirements (Dynamic Video Attributes)
- subject\_action\_requirements: gesture speed, behavior, emotional expression.
- camera\_action\_requirements: zoom, pan, tilt, dolly, handheld motion, stabilization quality.

format\_requirements
- video\_ratio
- resolution
- temporal consistency (frame-to-frame coherence)

cinematic\_grammar (Expanded Camera \& Film Language)
- shot\_size: e.g., wide, medium, close-up, extreme close-up.
- camera\_height: low angle, high angle, eye level.
- camera\_perspective: POV, objective, over-the-shoulder, long-shot, telephoto compression.
- camera\_angle: Dutch angle, frontal, profile, three-quarter angle.
- camera\_focus: shallow depth of field, deep focus.
- motion\_and\_speed: static, steady, tracking, crane, handheld wobble.
- shooting\_techniques: rack focus, bokeh, soft diffusion glow, slow shutter trails, motion blur.
- environment interaction: fog scattering, specular highlights, rim lights, volumetric lighting.
- compositional\_rules: rule of thirds, symmetry, leading lines.

---

Final Objective

Your task is to conduct a comprehensive video quality evaluation of the GeneratedVideo, referencing the OriginalVideo, ReferenceImages, and EditInstruction.
After providing feedback, you must also indicate how each issue impacts the overall generation quality (if no issues appear, omit this part).
You must carefully examine all six dimensions below:

Evaluation Checklist

1. Content Fidelity

(1) Subject:
    - Does the GeneratedVideo maintain high consistency with the OriginalVideo in terms of subject quantity, appearance, clothing, and details?
    - You must also consider the EditInstruction and ReferenceImages. If the instruction explicitly requires subject changes (e.g., "replace person A with person B from ReferenceImage1"), then the subject should reflect those changes rather than match OriginalVideo.
    - Are the identities of subjects maintained without abrupt changes or replacements (unless instructed)?
    - Are there any extra or missing key subjects (unless instructed)?

(2) Background:
    - Does the GeneratedVideo faithfully reproduce the time, location, and environmental setup of the OriginalVideo?
    - Are architectural layout structure, lighting effects, and environmental atmosphere preserved (unless EditInstruction requires background changes)?
    - If EditInstruction or ReferenceImages specify background modifications, evaluate whether those changes are correctly implemented.

(3) Events and Logic:
    - Does the GeneratedVideo preserve the core events and narrative structure of the OriginalVideo? Is the flow natural and coherent?

2. Style Consistency \& Visual Alignment

(1) Color \& Tone:
    - Does the GeneratedVideo match the OriginalVideo in overall\_tone, saturation, and contrast?
    - If ReferenceImages provide specific color palettes, are those colors accurately reflected in GeneratedVideo?

(2) Lighting \& Atmosphere:
    - Are lighting\_direction, lighting\_effect, and overall brightness consistent with the OriginalVideo's lighting layout and atmosphere?
    - If EditInstruction or ReferenceImages specify lighting changes, evaluate whether those changes are correctly applied.

(3) Image Style:
    - Is the image\_style consistent with the OriginalVideo (e.g., realistic, anime, cinematic)? Are there visually inconsistent or stylistically abrupt segments?
    - If EditInstruction requests a style change, evaluate whether the new style is correctly applied throughout.

3. Temporal \& Motion Coherence

In multi-reference image editing scenarios, the primary goal is successful edit execution. You should balance two aspects:
(1) Edit Success: Whether the edit specified in EditInstruction is correctly implemented
(2) Temporal Consistency: Whether appropriate consistency is maintained

Both aspects matter, but successful edit execution takes precedence. If an edit naturally requires motion or spatial relationship changes, those changes should not be penalized.

(1) Subject Actions:
    - Primary evaluation: Are the edits related to subject actions (if any) successfully executed according to EditInstruction?
    - Secondary evaluation: When EditInstruction does not explicitly modify actions, are the subject\_action\_requirements accurately and smoothly reproduced in GeneratedVideo? Are the action scale and rhythm consistent with OriginalVideo?

(2) Camera Actions:
    - Primary evaluation: Are the camera-related edits (if any) successfully implemented according to EditInstruction?
    - Secondary evaluation: When EditInstruction does not modify camera movement, are the camera\_action\_requirements preserved? Do zoom, pan, tilt, dolly, handheld motion, stabilization quality remain consistent with OriginalVideo?

(3) Relative Position:
    - Primary evaluation: Are spatial relationship edits (if any) correctly implemented according to EditInstruction?
    - Secondary evaluation: When EditInstruction does not modify spatial relationships, are the relative positions maintained? Note that certain edits (e.g., adding/removing objects, replacing subjects) may naturally alter these relationships and should not be penalized.

(4) Transitions \& Smoothness:
    - Are scene transitions and motion sequences smooth without flickering, stuttering, or unnatural jumps?
    - Do the main subjects and environment maintain smooth motion trajectories?

4. Technical Quality

(1) Generation Quality:
    - Does the GeneratedVideo contain artifacts, distortions, blurriness, or misalignments?
    - Are there issues like deformation of subjects or objects?

(2) Consistency:
    - Do the main subjects and objects maintain visual consistency across frames (appearance, clothing, features)?

5. Artistic Expressiveness \& Narrative Integrity

- Does the GeneratedVideo demonstrate artistic rhythm, lighting composition, and atmosphere?
- Is the narrative compelling with appropriate shot selection and scene transitions?

6. IP \& Privacy Compliance

- The GeneratedVideo must not contain recognizable copyrighted materials, including:
  * Brand logos, trademarks, or corporate IP
  * Licensed IP characters (e.g., Mickey Mouse, Marvel characters)
  * Recognizable public figures or celebrities (unless authorized)
  * Copyrighted artwork or designs
\end{tcolorbox}

\twocolumn
\end{document}